\documentclass[preprint,12pt]{IEEEtran}
\usepackage{cite}
\usepackage{float}
\usepackage{amsmath}    % For advanced math formatting
\usepackage{amssymb}    % For extra symbols
\usepackage{amsfonts}   % For \mathbb
\usepackage{mathrsfs}   % For calligraphic fonts
\usepackage{algorithmic}
\usepackage{graphicx}
\usepackage{textcomp}
\usepackage{makecell}
\usepackage{tabularx}
\usepackage{multirow}
\usepackage{bm}
\usepackage{booktabs}
\usepackage{hyperref}
\makeatletter
\AtBeginDocument{\DeclareMathVersion{bold}
\SetSymbolFont{operators}{bold}{T1}{times}{b}{n}
\SetMathAlphabet{\mathrm}{bold}{T1}{times}{b}{n}
\SetMathAlphabet{\mathit}{bold}{T1}{times}{b}{it}
\SetMathAlphabet{\mathbf}{bold}{T1}{times}{b}{n}
\SetMathAlphabet{\mathtt}{bold}{OT1}{pcr}{b}{n}
\SetSymbolFont{symbols}{bold}{OMS}{cmsy}{b}{n}
\renewcommand\boldmath{\@nomath\boldmath\mathversion{bold}}}
\makeatother

\def\BibTeX{{\rm B\kern-.05em{\sc i\kern-.025em b}\kern-.08em
    T\kern-.1667em\lower.7ex\hbox{E}\kern-.125emX}}

\begin{document}

\title{A Systematic Review of Open Datasets Used in Text-to-Image (T2I) Gen AI Model Safety}

\author{
    Trupti Bavalatti$^{1*}$, 
    Osama Ahmed$^{2*}$, 
    Dhaval Potdar$^{2*}$, 
    Rakeen Rouf$^{2*}$, 
    Faraz Jawed$^{2*}$, 
    Manish Kumar Govind$^{3}$, 
    and Siddharth Krishnan$^{3}$  
    \thanks{$^1$Meta, USA (e-mail: truptib@meta.com) and corresponding author}  
    \thanks{$^2$Duke University, USA (e-mail: \{osama.shawir, dhaval.potdar, rakeen.rouf, faraz.jawed\}@duke.edu)}  
    \thanks{$^3$University of North Carolina at Charlotte, USA (e-mail: skrishnan@uncc.edu)}  
    \thanks{*These authors contributed equally.}
}

\maketitle

\vspace{5pt}
\noindent\textbf{Author's Version:} This is the author's version of the paper.

\vspace{5pt}
\noindent\small\textcopyright~2025 IEEE. This work is licensed under the Creative Commons Attribution 4.0 International License (CC BY 4.0). For the definitive version, see  
\href{https://doi.org/10.1109/ACCESS.2025.3539933}{10.1109/ACCESS.2025.3539933}.  
\normalsize

\vspace{10pt}
\noindent\textit{\textbf{Disclaimer}: This research involves topics that may include disturbing results. Any explicit content has been redacted, and potentially disturbing results have been presented in a neutral and anonymized manner to minimize emotional distress to the readers.}

\vspace{10pt}
\begin{abstract}
Novel research aimed at text-to-image (T2I) generative AI safety often relies on publicly available datasets for training and evaluation, making the quality and composition of these datasets crucial. This paper presents a comprehensive review of the key datasets used in the T2I research, detailing their collection methods, compositions, semantic and syntactic diversity of prompts and the quality, coverage, and distribution of harm types in the datasets. By highlighting the strengths and limitations of the datasets, this study enables researchers to find the most relevant datasets for a use case, critically assess the downstream impacts of their work given the dataset distribution, particularly regarding model safety and ethical considerations, and also identify the gaps in dataset coverage and quality that future research may address.
\end{abstract}

\section{Introduction}
\label{sec:introduction}
\subsection{Overview of T2I Models and their Safety Stack}
The advancement of T2I generative AI models is disrupting traditional content creation methodologies, enabling users to generate highly photorealistic, imaginative, or personalized images from textual descriptions \cite{ramesh2021zeroshot}. Although their ability to generate visually compelling outputs based on simple textual prompts has caused their widespread adoption across industries \cite{saharia2022photorealistic}, the open-ended nature of these models poses challenges in ensuring that the generated outputs align with ethical, cultural, and societal norms \cite{bommasani2022opportunities}. The models can be misused for generating harmful or inappropriate content, including violent, explicit, or biased imagery as the models are trained on large-scale datasets that are composed of scraping content off the Internet and will inadvertently encode harmful patterns, stereotypes, or offensive content they learn from the training data \cite{prabhu2020large}. To safeguard end users and prevent misuse, an entire safety stack is built, composed of different techniques that are applied at different life-cycle stages of the model to make the generations safer, such as filtering out unsafe content from training data, fine-tuning the foundational model, or building a post-training mitigation layer consisting of prompt engineering and content moderation classifiers, such as input prompt classifiers, output image classifiers, or multimodal classifiers that can detect harmful prompt-image pairs \cite{openai2022dalle2}.

\subsection{Role of Labeled Datasets in Safety}
Labeled datasets, which consist of input text prompts and the corresponding generated image responses and annotations, whether they violate safety guidelines, are essential for building the components of the safety stack \cite{openai2022dalle2}. These datasets play a dual role: they are used to train/fine-tune the foundation model or train classifiers that detect and block harmful content, and they also serve as benchmarks for evaluating the safety performance of these classifiers or models. By capturing a diverse range of potential misuse cases, labeled datasets enable the creation of robust safety systems capable of mitigating the risks associated with T2I systems.

\subsection{Importance of Dataset Analysis in T2I Models}
Novel research on T2I safety is increasingly based on publicly available datasets; therefore, studying the composition of datasets that contain labeled prompt-to-image pairs is crucial for several reasons. These datasets are annotated to indicate whether the <prompt input, image output> combinations violate safety guidelines and serve as a core resource for training and evaluating the performance of these safety components. Understanding the composition ensures that the dataset adequately represents a wide range of harmful categories, including edge cases, which may challenge a model’s safety mechanisms. It helps to identify biases or gaps in coverage, such as over-representation of certain types of violations while under-representing others, which could result in unbalanced or ineffective classifiers. Analysis of the dataset composition will also reveal potential ambiguities or inconsistencies in labeling that may impact model training and evaluation \cite{hutchinson2021accountability}\cite{weidinger2021ethical}\cite{solaiman2021palms}. Moreover, a thorough understanding of the dataset helps researchers contextualize and appropriately interpret their findings, avoiding over-generalizations  or misjudgments about their model's safety performance \cite{bender2021stochastic}. This ensures that the conclusions drawn from the research are accurate and align with the limitations and strengths of the dataset.

\subsection{Objectives of the Research}
Although labeled datasets for T2I models exist, there are significant gaps in their systematic analysis. Many datasets lack detailed documentation regarding their composition, coverage across various harm types, diversity, topic representation, and labeling quality, making it challenging to assess their utility in safety applications. In addition, inconsistent labeling practices can undermine the effectiveness of content moderation systems trained on these datasets. This study aims to address these gaps by providing a comprehensive review of open-source labeled datasets used in T2I safety. By providing insights into the strengths and limitations of these datasets, the study provides researchers and practitioners with the knowledge needed to select appropriate datasets for a use case,  make accurate interpretations of their research findings, and design more effective safety interventions. Ultimately, this review aims to advance the development of robust, ethical, and safe T2I models.

\section{Related Work}
Datasets are critical to the performance and safety of generative models. Several studies have been done on the datasets used for training and finetuning LLMs. \cite{liu2024datasets} categorized large amount of datasets into pre-training corpora, instruction fine-tuning datasets, preference datasets, evaluation datasets and traditional NLP datasets and studied their challenges. More recently, a comprehensive study of preference datasets used in fine-tuning has been done in \cite{xiao2024preference} and a study of 16 pre-train datasets and 16 fine-tune datasets from qualitative perspective has been done in \cite{du2024survey}. There have also been systematic studies of LLM benchmarks itself, for instance \cite{li2024multimodal} surveys MLLM benchmarks, reviewing 211 benchmarks that assess MLLMs across understanding, reasoning, generation, and application and a detailed analysis of dataset constructions across diverse modalities has been provided and a critical assessment of 23 state-of-the-art LLM benchmarks using a unified evaluation framework through the lenses of people, process, and technology has been done in \cite{mcintosh2024inadequacies}.

For T2I applications, the research while not as extensive as on LLMs, is surely catching up. For safety, several studies have focused on analyzing the datasets used in T2I models and examining their biases, coverage, and diversity. Societal bias has shown to exist in datasets \cite{meister2023gender}\cite{wang2021revise}\cite{yang2020fairer} and a study of widely used datasets for text-to-image synthesis, including Oxford-102, CUB-200-2011, and COCO has been done in \cite{tan2023texttoimage},  calling inadequate diversity of real-world scenes. Similarly, a study of the LAION-400M dataset contains problematic content across misogyny, pornography, and malignant stereotypes \cite{birhane2021multimodal}. 

Most existing research is on analyzing certain datasets for specific harm types as shown above. There is a new body of work that is being done to address the gaps in the datasets. DataPerf introduced the Adversarial Nibbler challenge \cite{quaye2024nibbler}, emphasizing the importance of gathering prompts to identify harmful or biased outputs in T2I models. Other studies, such as LatentGuardCoPro \cite{liu2024latentguard}, have developed datasets specifically targeting unsafe input prompts, focusing on areas such as violence and hate speech.

By doing a systematic review of existing open datasets for T2I safety, we hope to encourage more such research to build comprehensive safety datasets covering broad classes of harm, that are composed of diverse set of topics and are that are in several languages. One such systematic study of safety datasets has been done for evaluating and improving safety of LLMs \cite{roettger2024safetyprompts}. Our research aims to do the same and fill the gap for T2I model safety. 

\section{Overview of Key Datasets}
In this study, we compiled the most prominent open-source datasets in the T2I model safety domain. Table 1 presents the number of prompts contained in each data source, with the majority coming from the "SafeGenExplicit56k" and "LatentGuardCoPro" datasets. The following is a brief description of each dataset.
\\
\\
\textbf{SafeGenExplicit56k:} The dataset comprises 56,000 textual prompts reflecting real-world instances of sexual exposure. The CLIP model \cite{radford2021clip} is utilized along with BLIP2 \cite{li2022blip} to generate multiple candidate text captions for a given explicit image. The optimal prompt is selected based on the highest CLIP score, ensuring the best alignment between the image and its textual description. \cite{li2024safegen}
\\
\\
\textbf{LatenGuardCoPro:} This dataset is a component of the Latent Guard framework \cite{liu2024latentguard}, aimed at improving safety in T2I models by identifying and blocking unsafe input prompts. The CoPro dataset includes prompts focused on harmful topics such as violence, hate, and sexual content. It was specifically developed for training and assessing the Latent Guard framework with data generated by selecting a range of unsafe concepts and using an LLM to create prompts associated with each concept.
\\
\\
\textbf{ART:}  The Meta Dataset (MD), a key component of the ART (Automatic Red-Teaming) \cite{li2024art} framework, comprises unique prompts paired with corresponding unsafe images collected from open-source prompt websites. This dataset is categorized into seven types of harmful content. The MD serves as the foundation for training the Guide Model and Writer Model within the ART framework, enabling the automatic generation and evaluation of safe prompts that could potentially trigger unsafe image generation from text-to-image models.
\\
\\
\textbf{P4D:} The P4D dataset \cite{chin2024prompting4debugging} was developed as part of a red-teaming framework aimed at identifying vulnerabilities in T2I diffusion models with safety mechanisms. The P4D framework utilizes prompt engineering techniques to generate problematic prompts that can bypass safety mechanisms in models, such as Safe Latent Diffusion  \cite{schramowski2023safe} and Erasing Stable Diffusion  \cite{erasingsd}.  This dataset includes prompts that lead to inappropriate outputs such as nudity, hate, or violent content, despite active safety filters. 
\\
\\
\textbf{Adversarial Nibbler:} This is a data-centric challenge, forming a part of the DataPerf challenge suite, and is organized with support from Kaggle and MLCommons. The objective of this crowdsourced challenge is to gather prompts to identify harmful or biased outputs from contemporary T2I models\cite{quaye2024nibbler}.
\\
\\
\textbf{MMA Diffusion:} This dataset is part of a framework designed to test and challenge the security of T2I models, particularly in generating inappropriate or Not-Safe-For-Work (NSFW) content. The text data generation process creates adversarial prompts that bypass safety filters, while maintaining the semantic intent of some author-selected NSFW prompts. The MMA Diffusion framework refines the adversarial prompts by substituting tokens and eliminating sensitive words \cite{mmadiffusion}
\\
\\
\textbf{SneakyPrompt:} This dataset is part of a framework for generating NSFW prompts that bypass T2I model safety filters. The data-generation process involves an automated attack framework that uses reinforcement learning to create adversarial prompts. Starting with a target prompt containing NSFW content, the system iteratively perturbs them until safety filters are bypassed while maintaining semantic similarity to the original prompt.\cite{yang2024sneakyprompt}
\\
\\
\textbf{I2P:} The Inappropriate Image Prompts (I2P) dataset, designed to evaluate image generation model safety, contains around 4,700 prompts covering various categories related to nudity, violence, and other potentially harmful content. It was developed as part of the Safe Latent Diffusion (SLD) project \cite{schramowski2023safe}. The prompts of the dataset originate from real-world data and focus on edge cases where models may generate unintended or explicit content, thereby providing a testbed for understanding model behavior and improving content safety measures \cite{schramowski2023safe}.
\\
\begin{table*}[h!]
\centering
\begin{tabular}{|r|l|l|l|}
\hline
\textbf{Dataset Name}& \textbf{\# Prompts} &\textbf{Data Source}&\textbf{GitHub License}\\
\hline
SafeGenExplicit56k \cite{li2024safegen}& 55,993 &Requested Access from paper authors  (xinfengli@zju.edu.cn)&Apache-2.0\\ \hline 
LatentGuardCoPro \cite{liu2024latentguard}& 46,072 &https://github.com/rt219/LatentGuard/tree/main/&BSD 3-Clause\\ \hline 
ART \cite{li2024art}& 6,571 &Requested Access from paper authors.  (kangjie001@e.ntu.edu.sg)&MIT\\ \hline 
P4D  \cite{chin2024prompting4debugging}& 4,467 &https://github.com/joycenerd/P4D/tree/main/data&MIT\\ \hline 
Adversarial Nibbler \cite{quaye2024nibbler}& 924 &https://github.com/google-research-datasets/adversarial-nibbler&Creative Commons\\ \hline 
MMA Diffusion \cite{mmadiffusion}& 812  &Request Access from authors of papers  (Yijun Yang1@cse.cuhk.edu.hk)&Creative Commons\\ \hline 
SneakyPrompt \cite{yang2024sneakyprompt}& 200  &https://github.com/Yuchen413/text2image\_safety/tree/main/data&MIT\\ \hline 
I2P \cite{schramowski2023safe}& 4,522 &https://huggingface.co/datasets/AIML-TUDA/i2p&MIT\\
\hline
\textbf{Total} & \textbf{119,561} &&\\
\hline
\end{tabular}
\vspace{5pt}
\\\caption{This table shows the total prompt count, dataset source and associated GitHub repository licenses by each dataset used in this study.}
\label{table:1}
\end{table*}

\section{Methodology}
\subsection{Prompt Annotation for Harm Categorization}
The categories (used interchangeably with "concepts" and "labels") employed to classify the prompts in this document correspond to Level 2 (L2) of the AIR taxonomy \cite{air2024}. A detailed description of the AIR taxonomy and associated classification methodology used to tag the prompts is provided in the following section. The original tags/labels assigned by the author to the prompts were excluded from this study to ensure consistency across different datasets, particularly with regard to the categories of harm used to define the prompts.

\subsection{Diversity of concepts in the prompts}
The analysis of diversity at the level of concepts represented in the dataset was performed using the generated L2 labels using the methodology described above.

\subsection{Prompt length distribution}
Prompt lengths were investigated in one of our experiments, as shown in Figure \ref{fig:countsentimentdistribution} (left). In this experiment, we split the sentences by " " (space) and counted the number of resulting elements in the list. The number represents the number of words displayed per prompt.

\subsection{Prompt sentiments}
Another insight from our aggregate dataset pertains to the distribution of sentiment at the prompt level, as shown in Figure \ref{fig:countsentimentdistribution} (right). In this experiment, we utilized the TextBlob \cite{loria2014textblobgithub} library in Python to calculate the sentiment scores for each prompt in the aggregated dataset. TextBlob's sentiment score ranges from -1 to 1, with -1 indicating very negative sentiment and 1 indicating very positive sentiment.

\subsection{Lexical and Syntactic Patterns in Prompts}
The most frequent words and bi-grams shown in Figures \ref{fig:frequniquedatasetword}, \ref{fig:frequniquedatasetbigram}, \ref{fig:frequniquecatword}, and \ref{fig:frequniquecatbigram} were determined through the following method: when a word appears as the most frequent across multiple datasets (or L2 categories), the subsequent most frequent word is iteratively selected for the remaining datasets (or categories). A similar approach was applied to the identification of bi-grams.

\subsection{Language diversity}
The languages were detected using a pre-trained model from the Hugging-Face Model Hub " papluca/xlm-roberta-base-language-detection \cite{papluca2021xlmroberta}.
\\

\subsection{Syntactic Diversity of Datasets}
Syntactic diversity looks at the variety in linguistic structure. The patterns of words, phrases, and sentences show how often different sequences of words or structures appear. A higher diversity indicates fewer repetitions and a broader range of language forms \cite{paraamr}. We measure syntactic diversity by calculating n-gram distinctness within a dataset in two ways: through intra-distinctness and inter-distinctness n-gram scores. The use of n-grams as a measure of diversity is further discussed by Dušek et al. \cite{dusek2020evaluating}.  
\\
\\
\textbf{Intra-distinctness n-gram scores} measure the diversity of n-grams within individual prompts. This approach evaluates each prompt separately and calculates the proportion of unique n-grams in the prompt relative to the total number of n-grams present. The intra-distinctness score provides insight into the syntactic variability of individual prompts, indicating how varied language patterns are within single prompts. Examining the distribution of intra-distinctness scores across all prompts allows us to see how diverse prompts are, on average, within a dataset. Figure~\ref{fig:agg_intra_scores} illustrates these scores for the aggregated dataset, whereas Figures ~\ref{fig:cat_intra_scores} and ~\ref{fig:source_intra_scores} provide detailed insights into the intra-distinctness scores across different harm categories and data sources, respectively.
\\
\\
\textbf{Inter-distinctness n-gram scores}, on the other hand, evaluate n-gram diversity across an entire dataset or subset of prompts. By examining the proportion of unique n-grams found throughout the dataset, interdistinctness provides a more robust measure of syntactic diversity at a broader level. Because inter-distinctness aggregates n-grams over the entire dataset, it offers insights into common or shared linguistic patterns, thus providing a more stable measure of syntactic diversity across diverse content. This is often more representative of overall diversity because it includes a larger pool of n-grams across different prompts, minimizing the biases that may arise from the structure of individual prompts. Table~\ref{tab:inter_scores} summarizes the inter-distinctness scores for the aggregated dataset as well as for different harm categories and data sources.
\\
\\
The intra- and inter-distinctness n-gram scores can be formalized as follows:

\begin{equation} \label{eqn:Intra_scores}
\text{Intra-Distinctness} = \frac{N_{\text{unique}} + \epsilon_{\text{nom}}}{C + \epsilon_{\text{den}}}
\end{equation}

\begin{equation} \label{eqn:Inter_scores}
\text{Inter-Distinctness} = \frac{N_{\text{global\_unique}} + \epsilon_{\text{nom}}}{C_{\text{total}} + \epsilon_{\text{den}}}
\end{equation}
\\
where:
\\
\begin{itemize}
    \item $N_{\text{unique}}$: Number of unique n-grams within a single prompt.
    \item $C$: Total number of n-grams within the same prompt.
    \item $N_{\text{global\_unique}}$: Number of unique n-grams across all prompts in a dataset.
    \item $C_{\text{total}}$: Total number of n-grams across all prompts.
    \item $\epsilon_{\text{nom}}$: A nominal epsilon value for numerical stability in the numerator (e.g., $10^{-12}$).
    \item $\epsilon_{\text{den}}$: A nominal epsilon value for numerical stability in the denominator (e.g., $10^{-5}$).
\end{itemize}
Inter-distinctness scores, calculated across entire datasets or larger subsets, are generally more robust in indicating syntactic diversity, as they consider the entire range of linguistic variability in the dataset, captured by the denominator, $C_{\text{total}}$, representing all n-grams. By contrast, intra-distinctness scores have the advantage of offering a granular view, revealing prompt-level syntactic variety, which can be plotted as distributions to better understand diversity across different subsets.

\subsection{Semantic Diversity of Datasets} 
Semantic Diversity examines the range of meanings of prompts. By comparing meanings using language model embeddings, we can see how conceptually distinct or related prompts are. Higher diversity means a wider spectrum of ideas, topics, and themes \cite{cevoli2021semantic}.
\\
\\
We measure semantic diversity \cite{hoffman2012semantic} using prompt embeddings and cosine distance metrics to capture the conceptual distinctness between prompts by analyzing their semantic relationships in a high-dimensional embedding space.
\\
\\
\textbf{Embedding Generation}
For embedding generation, we utilized the all-MiniLM-L6-v2 model, which is a lightweight yet effective sentence transformer that produces 384-dimensional dense vector representations of text \cite{reimers2019sentence}. This model is particularly suitable for our task because of its computational efficiency, while maintaining robust performance in semantic similarity tasks. The model demonstrates a strong capability in capturing semantic relationships in short text sequences while showing resilience to syntactic variations, making it ideal for analyzing the semantic content of text-to-image prompts.
\\
\\
\textbf{Cosine Distance Calculation}
We measure semantic diversity using the cosine distance derived from cosine similarity \cite{singhal2001modern}. For two prompt embeddings, $\mathbf{a}$ and $\mathbf{b}$, the cosine similarity is defined as
\begin{equation} \label{eqn:cossim}
\text{CosineSimilarity}(\mathbf{a}, \mathbf{b}) = \frac{\mathbf{a} \cdot \mathbf{b}}{\|\mathbf{a}\| \|\mathbf{b}\|}
\end{equation}
\\
The cosine distance is then calculated as follows:
\begin{equation} \label{eqn:cosdist}
\text{CosineDistance}(\mathbf{a}, \mathbf{b}) = 1 - \text{CosineSimilarity}(\mathbf{a}, \mathbf{b})
\end{equation}

\subsection{Coverage of harm types in the datasets}
\begin{figure*} 
    \centering
    \includegraphics[width=1\linewidth]{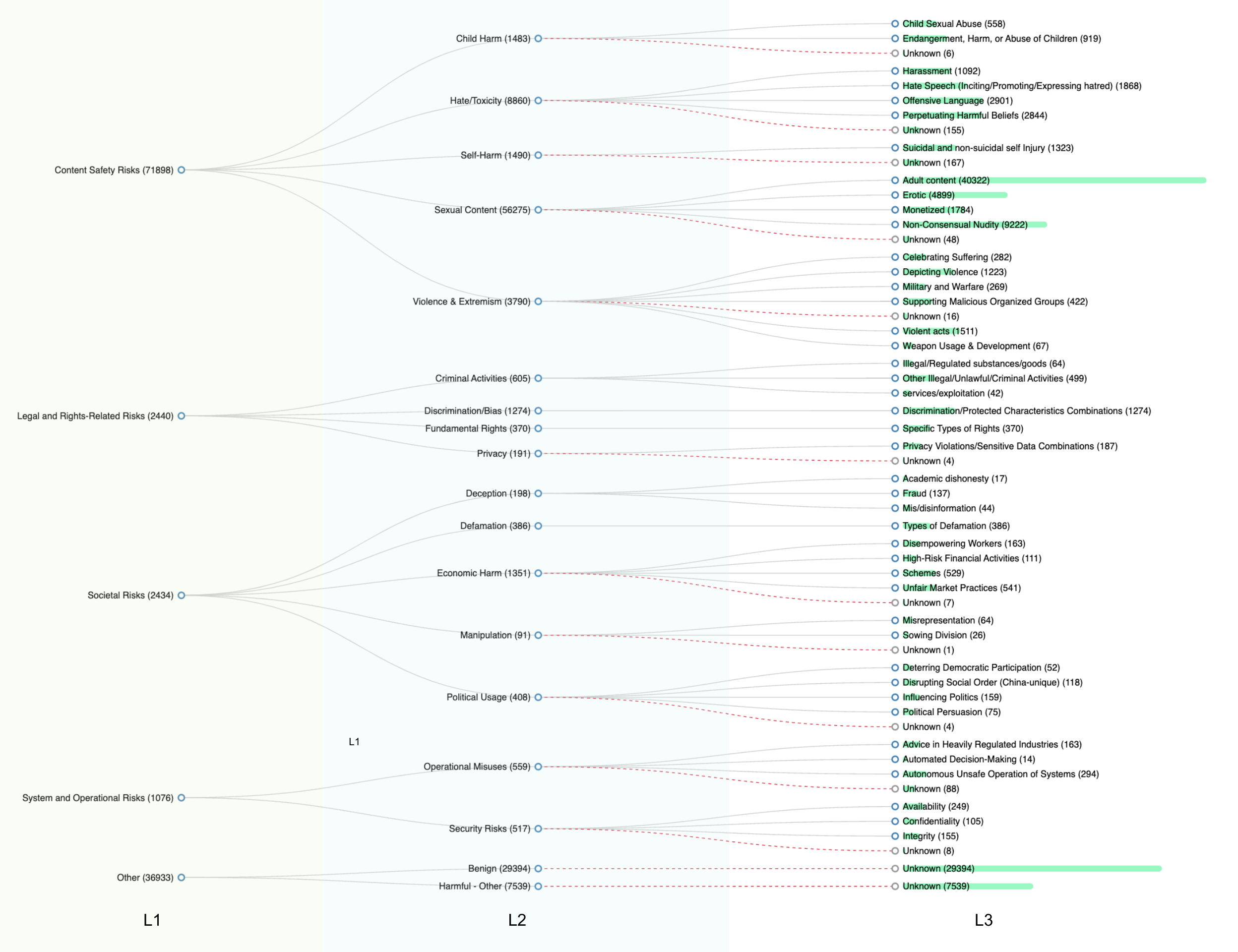}
    \caption{The complete dataset mapped onto AIR categories. 'Unknown' refers to prompts that were not mapped onto any of the categories in AIR.}
    \label{fig:fig_full}
\end{figure*}

\begin{figure}[t!]
    \centering
    \includegraphics[scale=0.3]{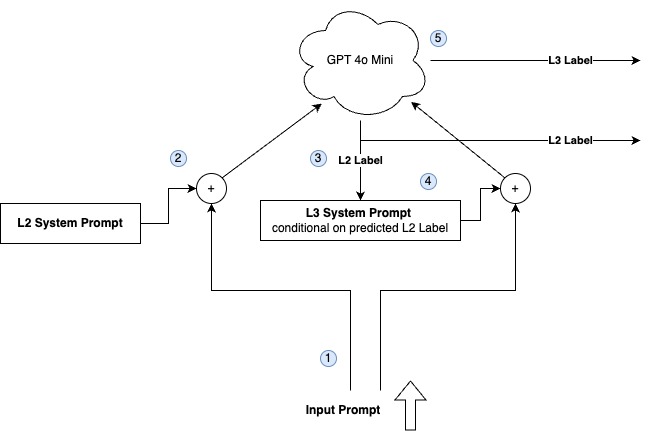}  % Scale down the image
    \caption{\textbf{The figure outlines the steps that an input prompt goes through in generating the L2 and L3 labels.}}
    \label{fig:airflowchart}
\end{figure}

The prompts in our consolidated dataset include author-assigned harm labels. However, these labels are often inconsistently defined, with terms like “NSFW” or “sexual content” being used interchangeably to represent similar harm categories. Because we wanted to study the coverage and composition of the datasets, we decided to standardize the labeling across all datasets and use a consistent method using the AIR (AI Risk) 2024 taxonomy \cite{air2024}. The AIR taxonomy categorizes AI risks into a structured framework based on analyses of government regulations \cite{aiact}\cite{execorder} and corporate policies. This taxonomy organizes 314 unique risks into a four-tier hierarchy. For this study, we use Level 2 (L2) and Level 3 (L3) of the hierarchy, with 16 and 45 categories, respectively, because their level of granularity aligns closely with the original author-provided labels in our dataset. 
\\
\\
In this study, we refer to a “harmful” prompt as any prompt intended to generate images that could cause harm, as defined by any category in the AIR-2024 taxonomy. Additionally, “harmful” prompts are those readily identifiable by a human as potentially harmful; they exclude prompts that lack explicit harmful intent or are too obfuscated to be easily recognized as harmful or are completely benign but were intended to test implicit model bias, stereotypes or ability to generate harmful images even when not explicitly prompted to. To address such ambiguous cases - and to include prompts designed by the authors to act as seed prompts that are subsequently converted into harmful variants - we introduce a “benign” L2 category. Finally, we also introduce a "harmful - other" L2 category that captures content that has identifiable potential for harm or disruption but does not clearly align with established risk categories. Thus, the total for L2 is 18 classes of harm. Because the two additional classes we introduce do not have corresponding L3 subcategories in the AIR taxonomy, we introduce a "Not Applicable" L3 label for such prompts. For completeness, we labeled the prompts L1 categories. Note that we do not generate predictions for L1 because they are too broad to provide any significant challenges for our classification algorithm. We simply map the predicted L2 categories to the corresponding parent L1 categories according to AIR taxonomy. The complete dataset mapped onto the AIR taxonomy is shown in Figure \ref{fig:fig_full}. The methodology for generating Level 2 and Level 3 labels is illustrated in Figure \ref{fig:airflowchart}.

\begin{itemize}
    \item \textbf{Steps 1 and 2:} A prompt is paired with a fixed system prompt that outlines the definitions of the 18 L2 categories along with 3 examples for each category. This query is sent to the GPT 4o Mini API.
    \item \textbf{Steps 3, 4, and 5:} As in step 3, the generated L2 label is saved at the output. It is also used to build the L3 system prompt. The L3 system prompt includes definitions and examples of only those L3 categories that correspond to a given L2 category. Note that the correct tagging at L3 depends on whether the L2 label is generated correctly. The query is sent to the GPT 4o Mini API again, and the L3 label is generated as the output in step 5.
\end{itemize}
\textbf{Confidence intervals of the AIR Labeling Method}: To evaluate our labeling method in a statistically robust manner, we hand-labeled 1,000 prompts from the consolidated dataset for L2 categories. We performed stratified sampling to ensure that the proportions of each category in the sample matched those of the population. However, this approach resulted in too few samples for rarer categories. To address this, we oversampled rarer categories to ensure at least 30 prompts per category. This adjustment increased the sample size to 1,316 prompts. Each prompt was then hand-labeled as "Correct" or "Incorrect" based on whether the label assigned by ChatGPT 4o-mini aligned with the definition provided in the L2 system prompt. 
\\
\\
We could not conduct a similar analysis for L3 categories. Achieving at least 30 samples per category while maintaining sample proportions consistent with the population would require a large number of prompts to be labeled manually. Moreover, because L3 labeling depends on the correct assignment of the parent L2 category, the available pool of prompts for L3 categories was further reduced, making it even more challenging to obtain the minimum number of labels required per category.
\\
\\
To estimate our confidence in L2 labels, we use a Bayesian approach \cite{bayesianCI}. Specifically, we assume a uniform prior distribution for the proportion of correctly labeled instances for each harm class, represented by a \(\text{Beta}(1, 1)\) prior. Coupled with a binomial sampling model to label the category correctly, the posterior distribution is modeled using a Beta distribution, \(\text{Beta}(\alpha, \beta)\), where \(\alpha = \text{successes} + 1\) and \(\beta = \text{failures} + 1\). Here, successes are the number of correctly labeled instances, and failures are incorrectly labeled instances. The posterior mean, which serves as the Bayesian point estimate for the proportion, is calculated as
\\
\[
\text{Posterior Mean} = \frac{\alpha}{\alpha + \beta}.
\]
\\
We report the 95\% Credible Intervals in Table \ref{tab:l2_accuracy_intervals}.

\subsection{Prompt Tagging and Thematic Analysis}
To better understand the thematic composition of prompts within each harm category, we implemented an advanced tagging system using GPT-4o. This system was designed to identify key themes, topics, and semantic patterns within prompts, providing a granular view of content distribution across AIR taxonomy categories. Our approach was inspired by the work of Lu et al. on InsTag \cite{instag}, who explored instruction tagging for analyzing supervised fine-tuning of large language models and adapted similar principles to suit the context of thematic prompt analysis.
\\
\\
The tagging system was designed to generate three to five specific and descriptive tags for each prompt, ensuring consistency across various harm categories while addressing ethical and safety considerations. It captures both the surface-level content and deeper implicit implications to provide a comprehensive understanding of each prompt.
\\
\\
To achieve this, we implemented a specialized prompt engineering framework (reached after several iterations) that enabled the language model to evaluate prompts from multiple perspectives. This framework analyzes four key dimensions.
\begin{enumerate}
    \item \textbf{Content Analysis:} Identification of primary themes and subject matter
    \item \textbf{Context Evaluation:} Assessment of situational factors
    \item \textbf{Implication Detection:} Recognition of potential consequences and ethical concerns
    \item \textbf{Domain Specificity:} Consideration of required knowledge or expertise
\end{enumerate}
Although explicitly mentioned in our system prompt to return tags, this methodology occasionally led to prompts being blocked by the model owing to safety and ethical guidelines. This was particularly evident in sensitive categories such as child harm and Violence and Extremism, where responses like "I'm sorry, I cannot do this" or "Inappropriate" were frequently generated, reflecting the model's adherence to ethical boundaries.

\section{Results}
\subsection{Prompt Annotation for Harm Categorization}
\begin{figure} 
    \centering
    \includegraphics[width=1\linewidth]{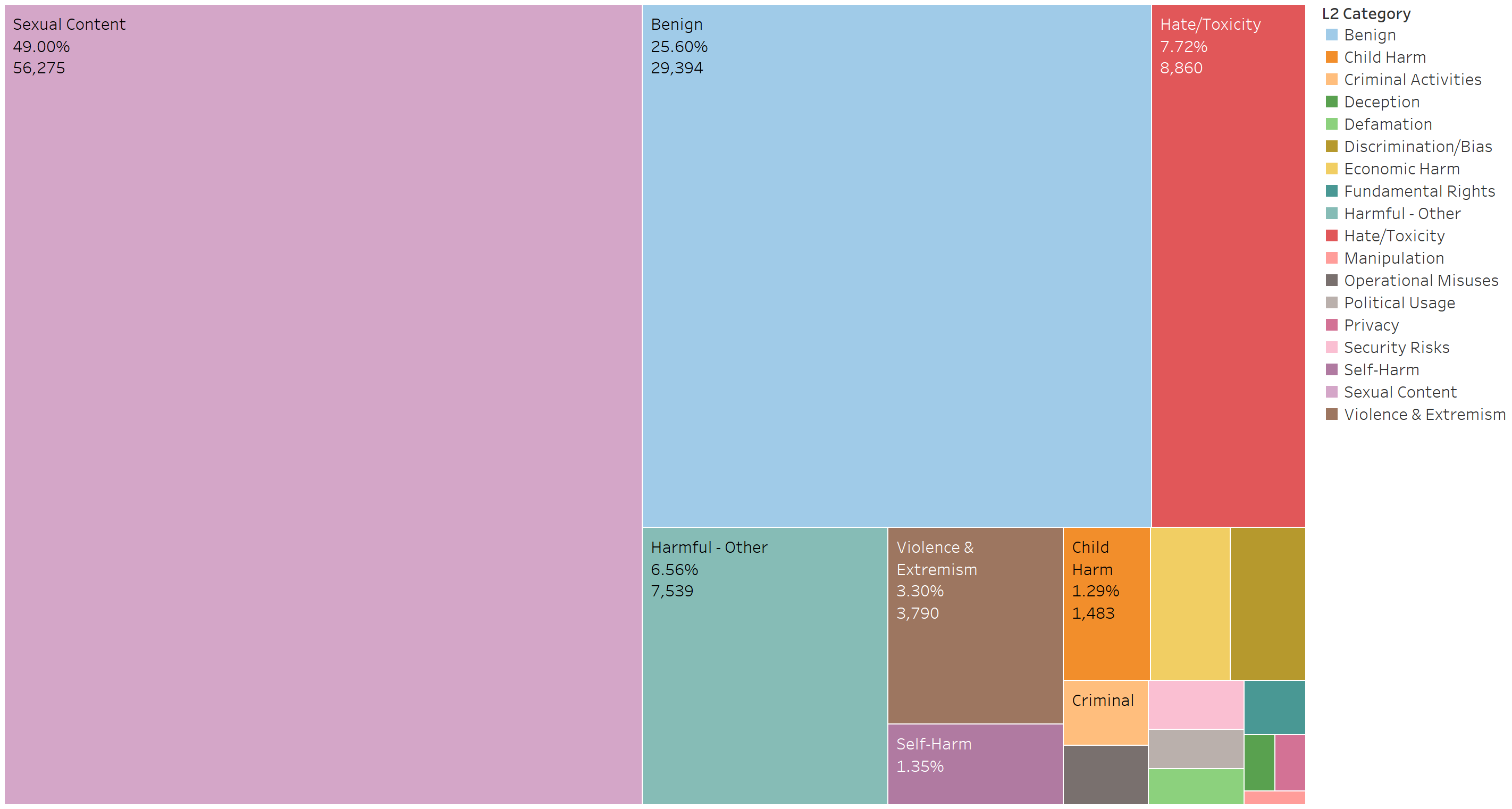}
    \caption{This figure shows the number of prompts by class of harmful concept found in our compiled dataset. The current prompt curation of the T2I space is very heavily focused on sexual content. Less focus is put on categories such as violence.}
    \label{fig:catpromptdist}
\end{figure}

\begin{figure*}
    \centering
    \includegraphics[width=1\linewidth]{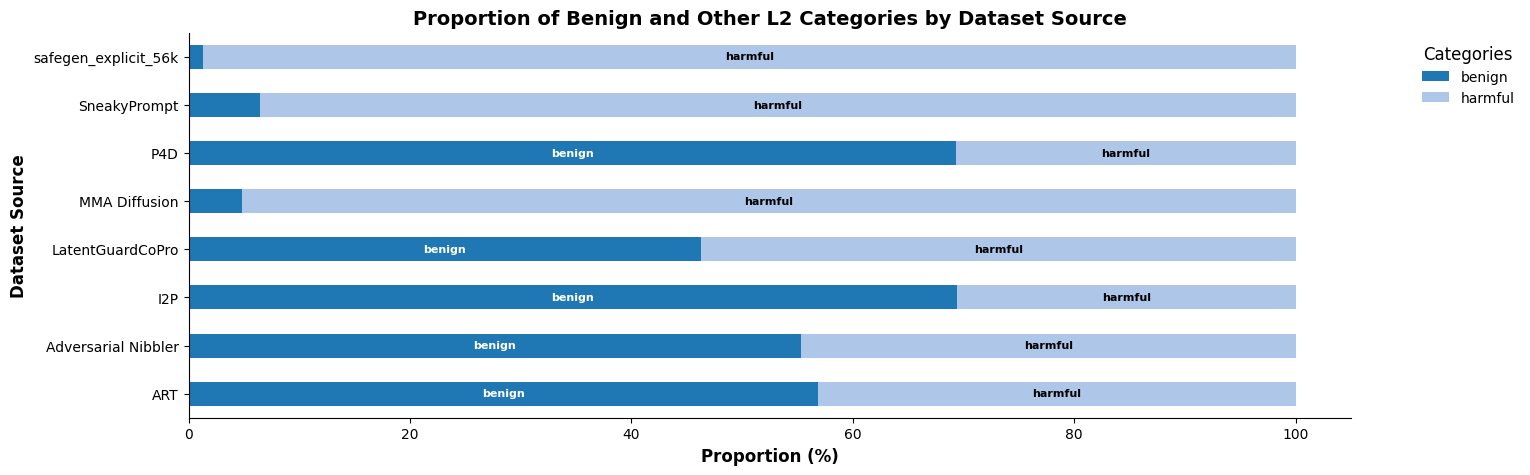}
    \caption{This figure shows the composition of each dataset by benign and harmful concepts. Most datasets contain a significant number of benign prompts. Benign prompts are defined as prompts classified as innocuous by our AIR classifier.  L2 categories are defined as the level 2 taxonomy/categories from the AIR taxonomy.}
    \label{fig:harmfultobenign}
\end{figure*}

Approximately 49 percent of all the prompts combined pertain to the concept of "Sexual Content, " with no other harmful category exceeding 7.72 percent. This indicates that open-source T2I safety prompts focus predominantly on sexual content. Although the generation of sexual content was a key area in early T2I model jailbreaks \cite{buolamwini2023unmasking}, it remains crucial to develop comprehensive datasets for other harmful concepts.  Our study reveals that important classes of harm have low coverage, such as discrimination against protected groups, Misinformation, Criminal Activities including Child Nudity, Trafficking, Social Issues, and Health Concerns. Another important observation is that approximately 25.6 percent of the prompts in our consolidated dataset are comprised of benign prompts. We define benign prompts as any prompts that do not explicitly ask for the generation of a harmful image. Figure \ref{fig:catpromptdist} illustrates the distribution of prompts according to the category of harmful content in our dataset.  
\\
\\
Upon examining the composition of harmful versus benign prompts across the datasets in Figure \ref{fig:harmfultobenign}, it is observed that the majority of datasets contain a substantial proportion (greater than 45 percent) of benign prompts. In contrast, the datasets "Safegen\_explicit\_56K", "SneakyPrompt,” and "MMA diffusion" stand out as the top three datasets in terms of containing prompts that are explicitly harmful. Notably, each of these datasets includes fewer than 10 percent benign prompts.

\subsection{Diversity of concepts in the prompts}
\begin{figure*}
    \centering
    \includegraphics[width=1\linewidth]{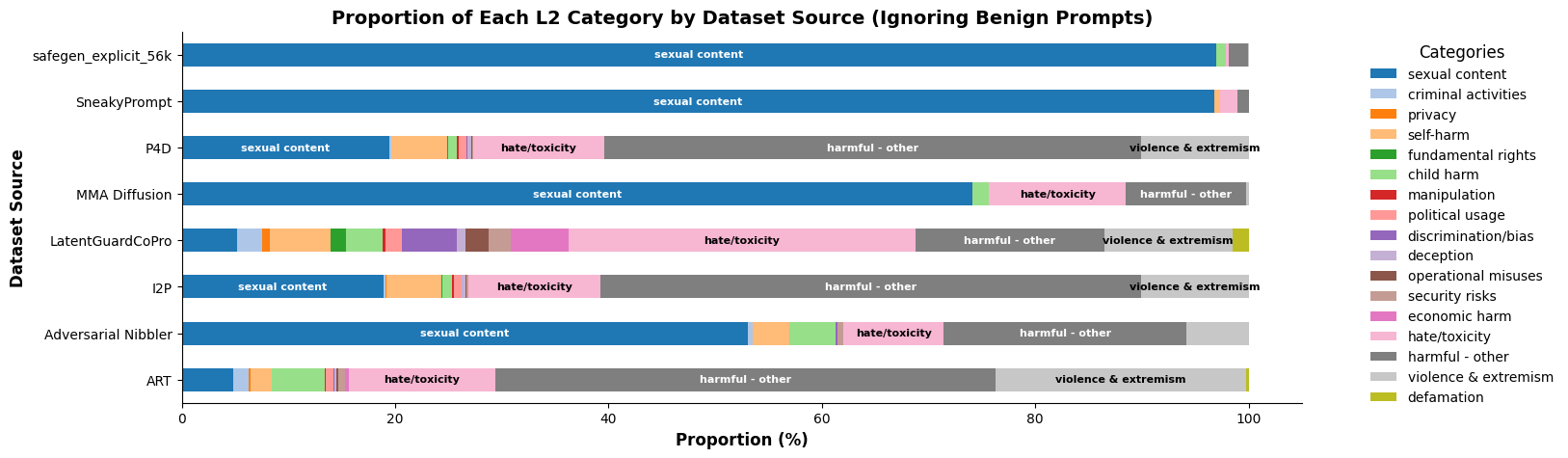}
    
    \caption{This figure shows the composition of each dataset by concept. Some of the most diverse datasets in our study were I2P, ART, P4D, and Latentguard CoPro. Diverse datasets are defined as those that have a balanced mix of harmful concepts. L2 categories are defined as the level 2 taxonomy/categories from the AIR taxonomy.}
    \label{fig:conceptcomposition}
\end{figure*}

When training models to detect harmful T2I outputs, it is important to ensure that a conceptually diverse dataset is used. Diverse harm concepts in training enhance the generalization capabilities and adaptability of the safety model to new or evolving threats. Figure \ref{fig:conceptcomposition} illustrates the conceptual composition of each curated dataset used in the study. The datasets "SafeGenExplicit56k,” "MMA Diffusion,” and "SneakyPrompt" were identified as having the poorest balance in terms of harm concept coverage. Conversely, the "P4D," "LatenGuardCoPro," and "ART" datasets exhibit the best balance. 

\subsection{Prompt length distribution}
\begin{figure*}
    \centering
    \includegraphics[width=1\linewidth]{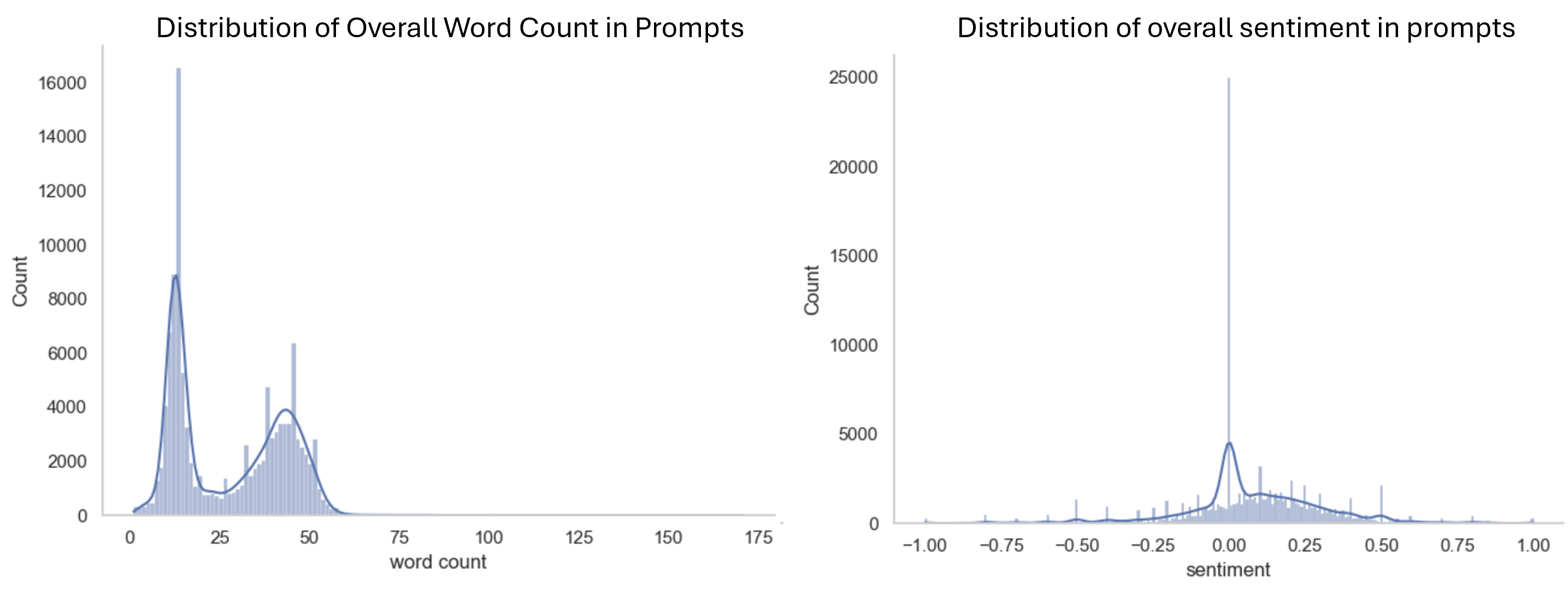}
    \caption{Salient distributions of the overall dataset. \textbf{Left: }The distribution of the lengths of the prompts exhibits a bimodal pattern, indicating two distinct peaks. Notably, most of these prompts are brief, typically under 50 words. \textbf{Right:} Even though we are dealing with unsafe T2I prompts, the overall sentiment distribution across all prompts is right-skewed toward positive. All categories of harm from the aggregated dataset exhibit similar sentiment distributions. }
    \label{fig:countsentimentdistribution}
\end{figure*}

Our consolidated dataset reveals a bimodal distribution of word count in our prompts, with peaks around 12 and 45 words per prompt. When we looked at the individual categories, we saw a unimodal distribution close to one of the previous peaks. Overall, this suggests that, in general, the prompts found in our consolidated dataset are brief. Therefore, there appears to be an opportunity to incorporate longer prompts into the open-source dataset space. Longer prompts may involve more complex or nuanced forms of harm that are difficult for T2I models to detect. Figure \ref{fig:countsentimentdistribution} (left) illustrates the distribution of prompt lengths in our collective dataset. Furthermore, examining individual harm categories reveals a unimodal distribution centered around either the first or second peak identified earlier.

\subsection{Prompt sentiments}
Although our dataset primarily consists of unsafe T2I prompts, the overall sentiment distribution is right-skewed (i.e., it has a pronounced right tail). This indicates that, on average, prompts are more likely to have positive sentiments than negative ones. Moreover, all the harm categories represented in our aggregated dataset (given sufficient samples) displayed a similar sentiment score distribution. This finding is somewhat surprising and challenges the assumption that harmful content is correlated with negative sentiments. Thus, there are doubts regarding the effectiveness of contemporary sentiment-based prompt filtering in T2I generative AI models.

\subsection{Lexical and Syntactic Patterns in Prompts}
\begin{figure*}
    \centering
    \includegraphics[width=1\linewidth]{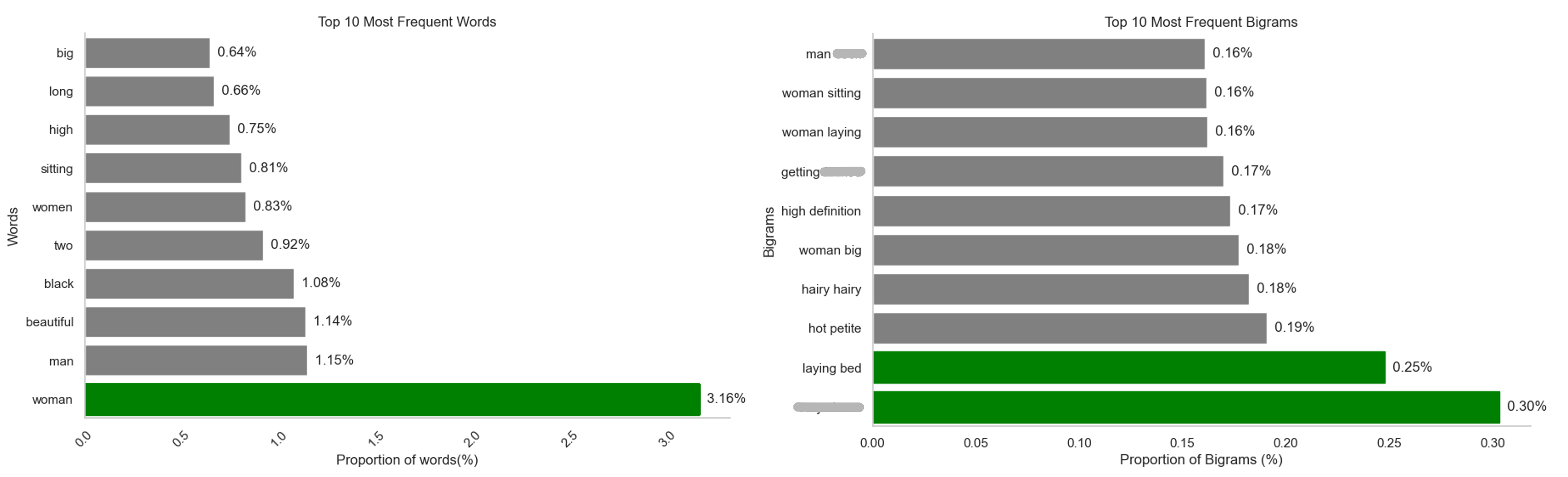}
    \caption{Left: This figure displays the most frequent words in the aggregated dataset. Notably, the word "woman" constitutes a staggering 3.16 percent of the entire dataset.  Right: This figure presents the most frequent bi-grams in the aggregated dataset.}
    \label{fig:frequent_ngrams}
\end{figure*}

\begin{figure}
    \centering
    \includegraphics[width=1\linewidth]{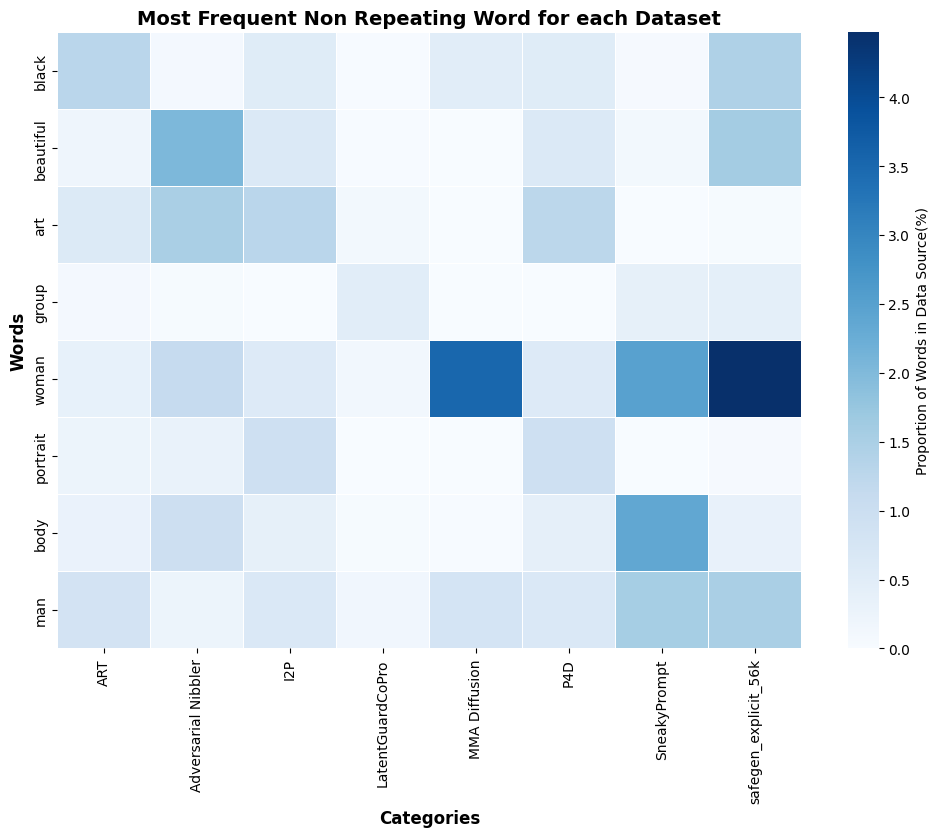}
    \caption{This heat map illustrates the most frequent non-repeating words for each of our datasets. In instances where a word appears to be the most frequent in multiple datasets, the next most frequent word is recursively selected for the subsequent datasets. The Y-axis displays the most frequent words, whereas the X-axis corresponds to the various datasets. Color intensity reflects the relative proportion of each word within its respective dataset, with darker shades indicating higher proportions. This visualization facilitates the identification of key terms associated with each dataset, highlighting how the most frequent words contribute disproportionately to the overall composition of each dataset.}
    \label{fig:frequniquedatasetword}
\end{figure}

\begin{figure}
    \centering
    \includegraphics[width=1\linewidth]{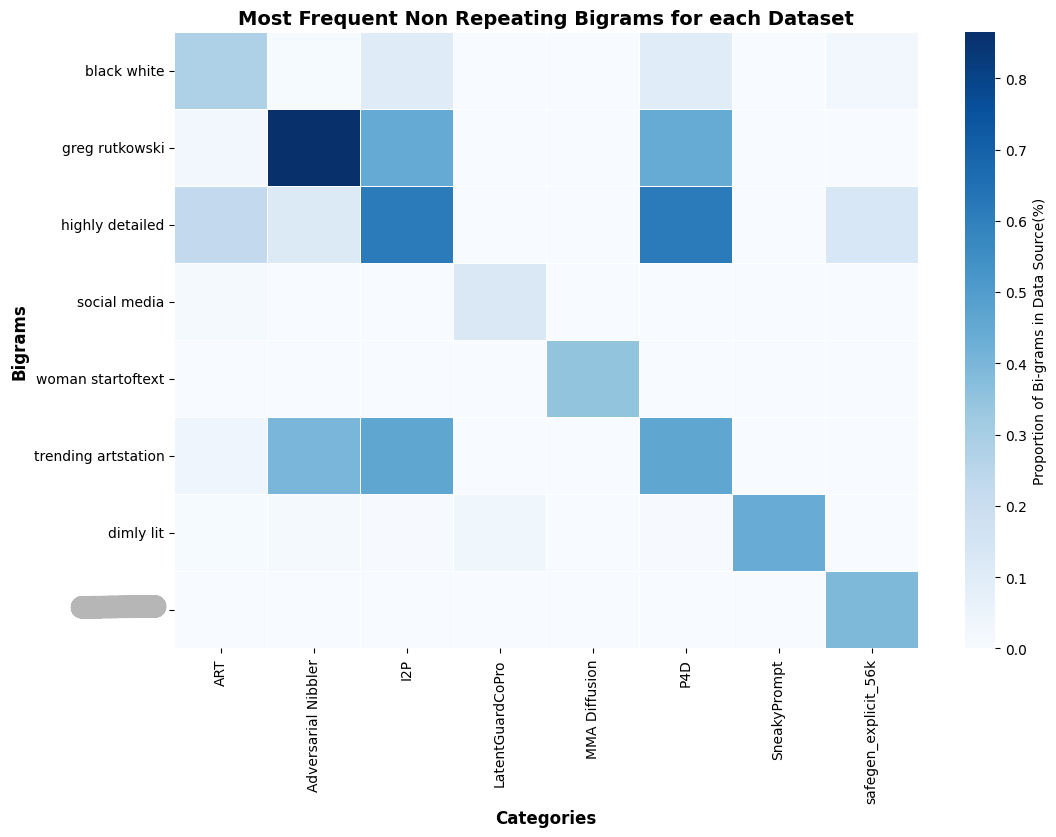}
    \caption{Similar to the previous figure, this heat map illustrates the most frequent non-repeating bi-grams for each of our datasets. This visualization facilitates the identification of key 2-word phrases associated with the dataset, highlighting how the most frequent bi-grams contribute disproportionately to the overall composition of each dataset.}
    \label{fig:frequniquedatasetbigram}
\end{figure}

\begin{figure}
    \centering
    \includegraphics[width=1\linewidth]{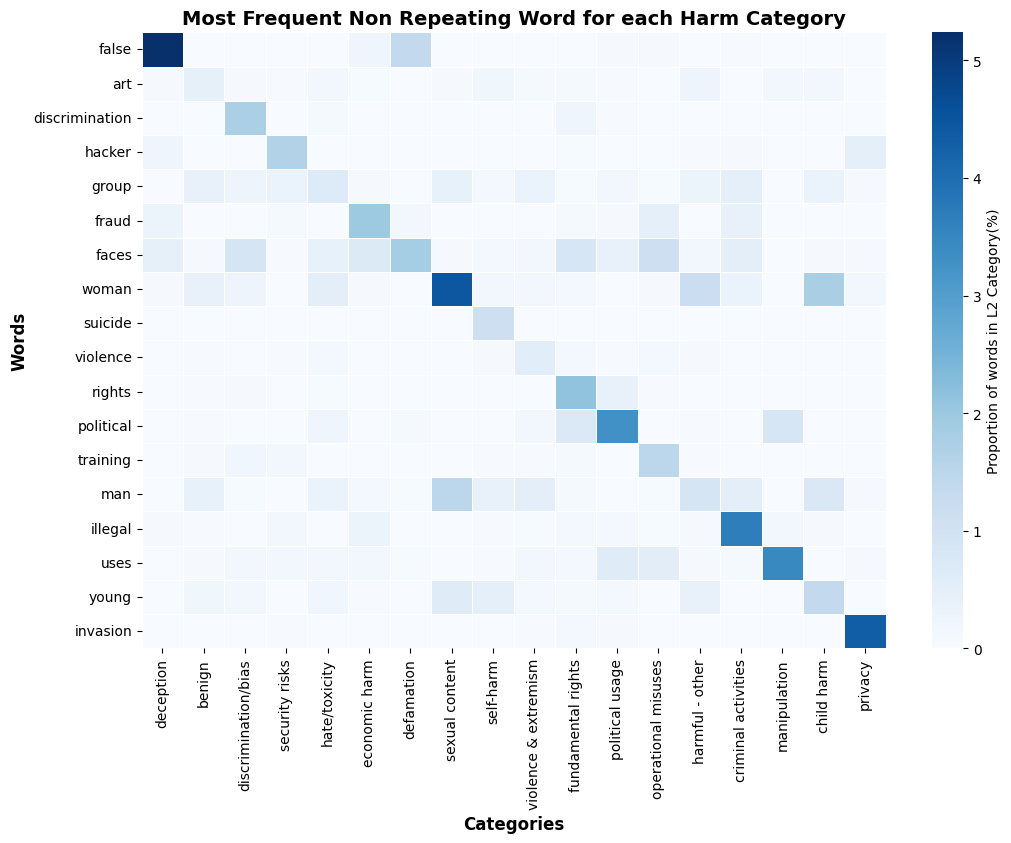}
    \caption{This heat map illustrates the most frequent non-repeating words for each L2 harm category. In instances where a word appears to be the most frequent in multiple categories, the next most frequent word is recursively selected for the subsequent categories. The Y-axis displays the most frequent words, while the X-axis corresponds to the various harm categories. Color intensity reflects the relative proportion of each word within its respective category, with darker shades indicating higher proportions. This visualization facilitates the identification of key terms associated with each harm category, highlighting how the most frequent words contribute disproportionately to the overall composition of each category.}
    \label{fig:frequniquecatword}
\end{figure}

\begin{figure}
    \centering
    \includegraphics[width=1\linewidth]{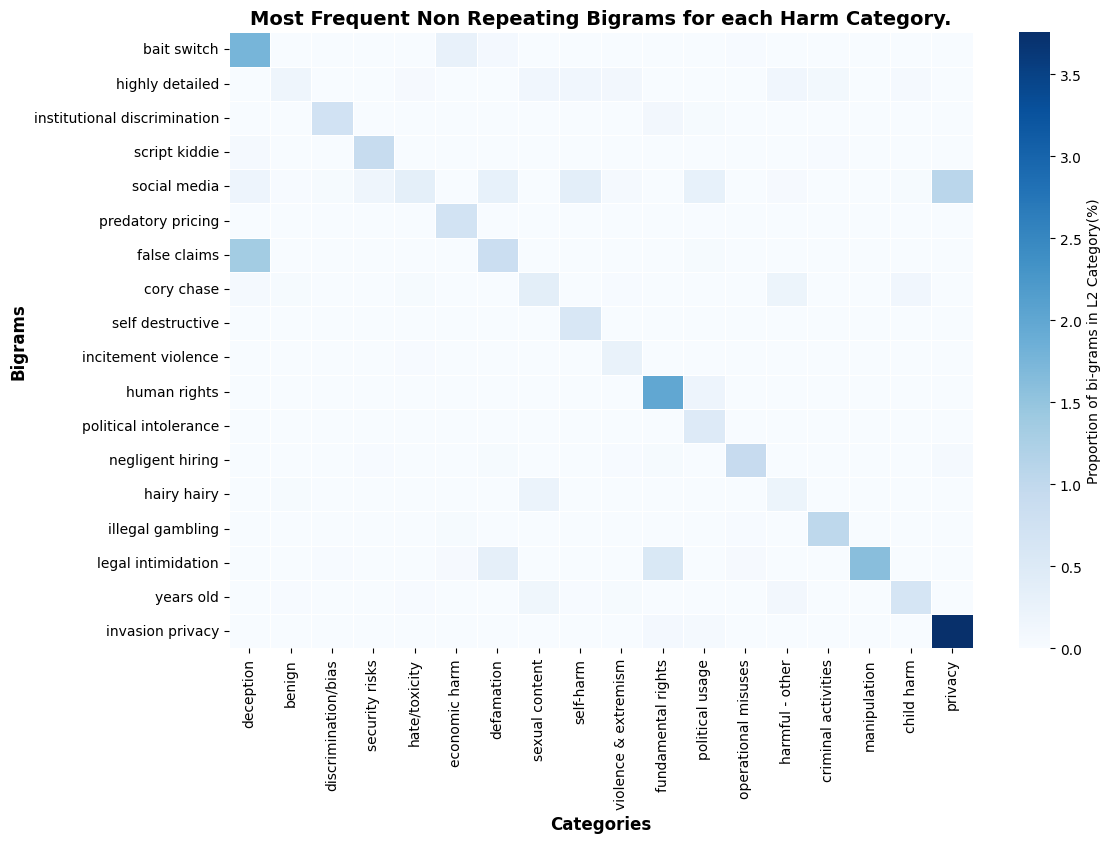}
    \caption{Similar to the last figure, this heat map illustrates the most frequent non-repeating bi-grams for each L2 harm category. This visualization facilitates the identification of two keyword phrases associated with each harm category, highlighting how the most frequent bi-grams contribute disproportionately to the overall composition of each category.
}
    \label{fig:frequniquecatbigram}
\end{figure}

The analysis of the most frequent words in the aggregated dataset reveals several noteworthy patterns, as illustrated in Figure \ref{fig:frequent_ngrams} (left). The word "woman" appears with remarkable frequency, constituting 3.16 percent of the entire dataset. This disproportionate representation is partially attributed to the inclusion of datasets focused on sexual content. Additionally, the top 10 most frequently used words accounted for 11.14 percent of all the words in the dataset, suggesting a significant lack of syntactic diversity within safety prompts. There is also a sharp decline in word frequency after the most common terms, indicating an uneven distribution of word usage across the dataset. The examination of bigrams (two-word sequences) in the dataset further reinforces the observations from the individual word analysis. The top 10 bi-grams frequencies are illustrated in Figure \ref{fig:frequent_ngrams} (right). Certain bi-grams, representing actor names (0.3 percent) and "laying bed" (0.25 percent), appear with notable frequency, suggesting a potential over-representation of specific scenarios or individuals in the dataset. The top 10 most frequent bi-grams constitute 1.92 percent of all bi-grams in the dataset, indicating a similar pattern of concentration as observed with individual words. As with individual words, there is a significant decrease in frequency after the most common bi-grams, further highlighting the uneven distribution of language patterns in the dataset. 
\\
\\
At a more granular level, Figure \ref{fig:frequniquedatasetword} presents the most frequent non-repeating words in the dataset. Recall that when a word is the most frequent across multiple datasets, the next most frequent word is recursively selected for the subsequent datasets. This approach reveals that terms like "woman," "body," "man," and "art" are disproportionately prominent in certain datasets. Similarly, Figure \ref{fig:frequniquedatasetbigram} shows the most frequent non-repeating bi-grams by the dataset. Bi-grams such as "highly detailed,", artist name (referring to lot of stylization prompts that explicitly ask "in the style of") and some actor names dominate their respective datasets, underscoring specific thematic biases.
\\
\\
Expanding the analysis to L2 harm categories, Figure \ref{fig:frequniquecatword} shows the most frequent non-repeating words. Terms like "woman," "false," and "use" appear disproportionately in some L2 categories, mirroring patterns observed at the dataset level. Similarly, Figure \ref{fig:frequniquecatbigram} examines the most common bi-grams within the L2 harm categories, highlighting the lack of diversity. Bi-grams such as "bait switch" and "invasion privacy" make up a significant portion of bi-grams within their respective categories, further emphasizing the concentration of language patterns within specific contexts.

\subsection{Language diversity}
Table 2 shows that most of our prompts are in English. Other languages made up only 1.6 percent of all prompts. This points to a lack of coverage of non-English prompts.
\begin{table}[h!] \centering \begin{tabular}{|l|r|} \hline \textbf{Language} & \textbf{Count} \\ \hline English & 113004 \\ Swahili & 636 \\ French & 267 \\ Italian & 204 \\ Portuguese & 137 \\ Dutch & 136 \\ Urdu & 117 \\ Hindi & 100 \\ Spanish & 85 \\ Polish & 66 \\ Turkish & 32 \\ German & 30 \\ Bulgarian & 18 \\ Russian & 6 \\ Arabic & 2 \\ Greek & 1 \\ \hline \end{tabular} 
\vspace{5pt}
\caption{Number of prompts by language as classified by  the "papluca/xlm-roberta-base-language-detection"  model, from Hugging Face Model Hub.}
\label{tab:my_table}

\end{table}

\subsection{Syntactic Diversity of Datasets}
\begin{figure}
    \centering
    \includegraphics[width=1\linewidth]{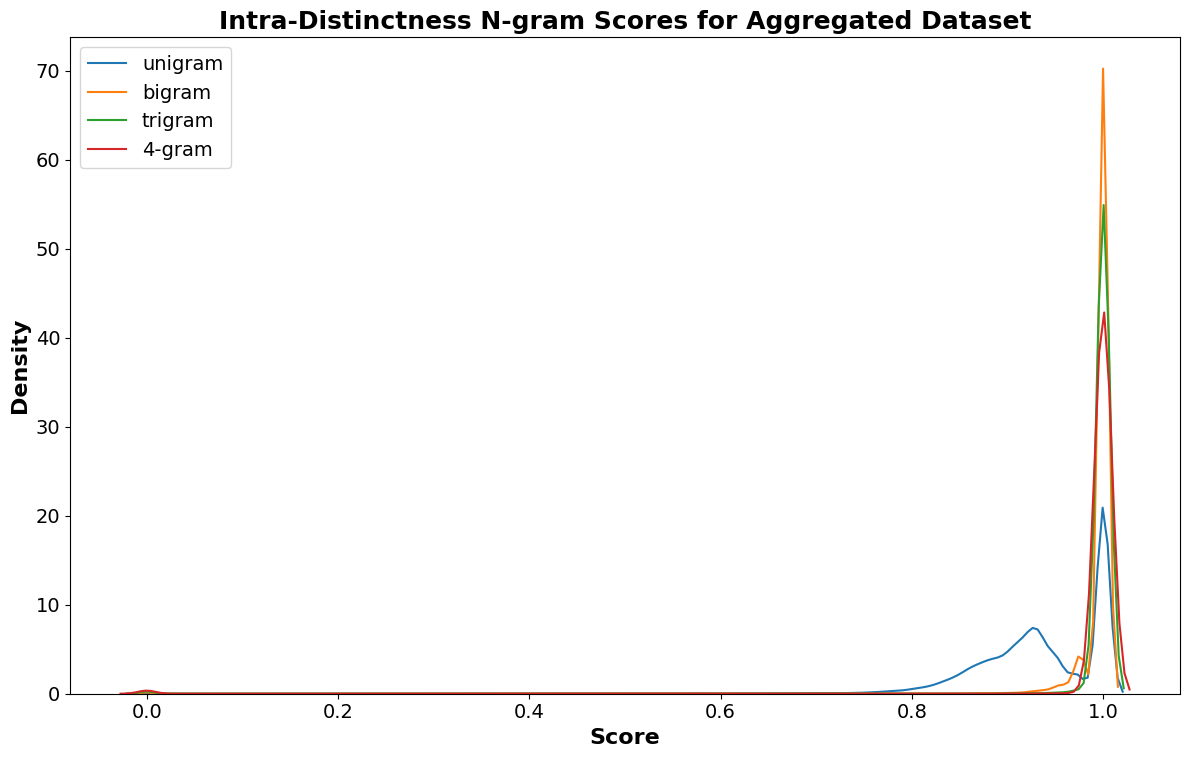}
    \caption{Intra-Distinctness N-gram Scores for Aggregated Dataset: This figure shows the density plots of intra-distinctness scores for un-igrams, bi-grams, trigrams, and 4-grams across the entire aggregated dataset. The scores indicate the diversity of n-grams within the dataset.}
    \label{fig:agg_intra_scores}
\end{figure}

\begin{figure}
    \centering
    \includegraphics[width=1\linewidth]{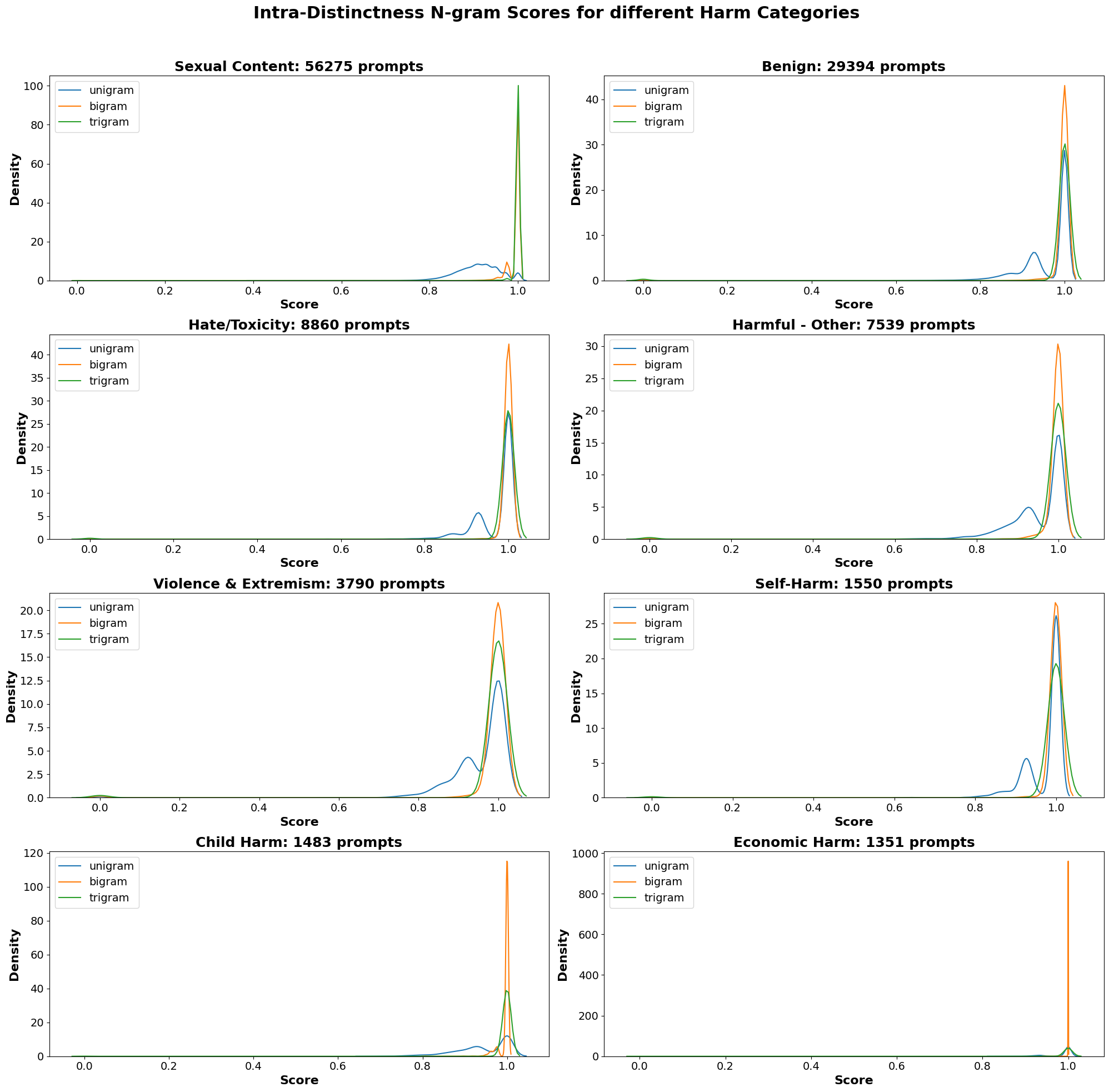}
    \caption{Intra-Distinctness N-gram Scores for Different Harm Categories: This grid of plots displays the density of intra-distinctness scores for uni-grams, bi-grams, and trigrams across various harm categories. This plot shows the scores for the 8 categories with the largest number of prompts.}
    \label{fig:cat_intra_scores}
\end{figure}

\begin{figure}
    \centering
    \includegraphics[width=1\linewidth]{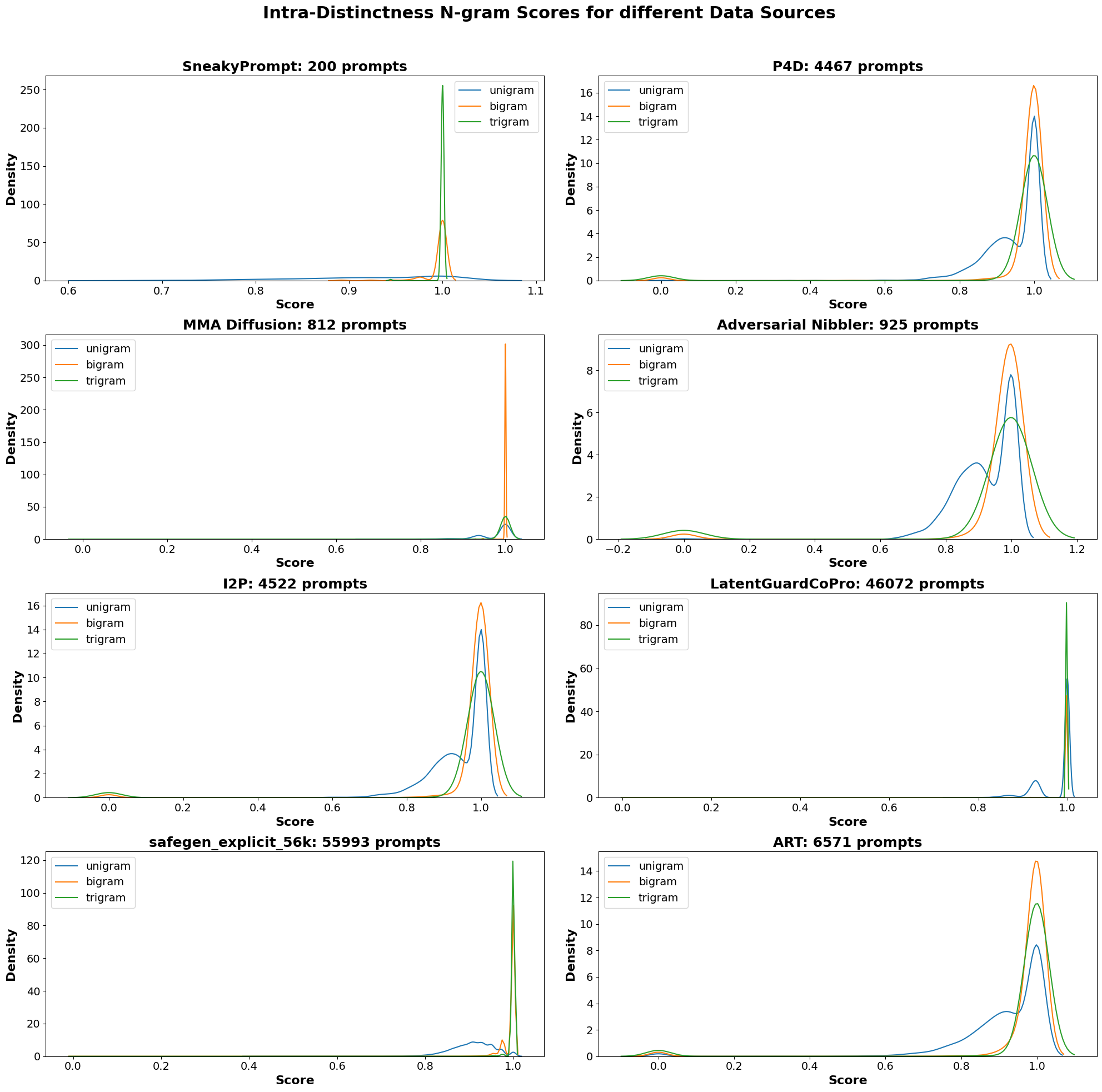}
    \caption{Intra-Distinctness N-gram Scores for Different Data Sources: This grid of plots displays the density of intra-distinctness scores for uni-grams, bi-grams, and trigrams across various data sources. Each subplot represents a different data source, showing the diversity of the n-grams within each source. }
    \label{fig:source_intra_scores}
\end{figure}

We examined the syntactic diversity of our aggregated dataset, which includes prompts from eight distinct data sources categorized into 16 harm categories. Table~\ref{tab:inter_scores} shows the inter-distinctness scores for the aggregated dataset and for each data source and harm category. Figures~\ref{fig:agg_intra_scores}, \ref{fig:cat_intra_scores}, and \ref{fig:source_intra_scores} show the distribution of intra-distinctness scores across all prompts organized by the area of harm and the data source. Owing to the wide variation in the number of prompts across different categories, we only plotted the intra-distinctness n-gram scores for the eight categories with the largest number of prompts. This is because the distribution of scores for categories with fewer prompts is too narrow to be visually informative.
\begin{table}[h!]
\centering
\begin{tabular}{|l|c|c|c|}
\hline
 \textbf{Data} & \makecell{\textbf{Unigram} \\ \textbf{Score}} & \makecell{\textbf{Bigram} \\ \textbf{Score}} & \makecell{\textbf{Trigram} \\ \textbf{Score}} \\
\hline
Aggregated Dataset & 0.015491 & 0.262871 & 0.510254 \\
\hline
Benign Category & 0.062376 & 0.399624 & 0.652069 \\
Child Harm Category & 0.154031 & 0.487391 & 0.654285 \\
Criminal Activities Category & 0.221182 & 0.581197 & 0.783951 \\
Deception Category & 0.355425 & 0.706038 & 0.830362 \\
Defamation Category & 0.287712 & 0.655423 & 0.819196 \\
Discrimination/Bias Category & 0.172094 & 0.566610 & 0.772992 \\
Economic Harm Category & 0.172937 & 0.550581 & 0.767606 \\
Fundamental Rights Category & 0.319803 & 0.717437 & 0.872715 \\
Harmful - Other Category & 0.120256 & 0.524997 & 0.735410 \\
Hate/Toxicity Category & 0.099855 & 0.455250 & 0.702390 \\
Manipulation Category & \textbf{0.444712} & \textbf{0.780679} & \textbf{0.887143} \\
Operational Misuses Category & 0.294225 & 0.696255 & 0.866724 \\
Political Usage Category & 0.341278 & 0.747194 & 0.894845 \\
Privacy Category & 0.338462 & 0.648114 & 0.789723 \\
Security Risks Category & 0.261566 & 0.641975 & 0.813803 \\
Self-Harm Category & 0.193194 & 0.587746 & 0.796296 \\
Sexual Content Category & 0.011692 & 0.249341 & 0.487181 \\
Violence \& Extremism Category & 0.145055 & 0.520511 & 0.725857 \\
\hline
ART Dataset & 0.086489 & 0.347865 & 0.484980 \\
Adversarial Nibbler Dataset & 0.125536 & 0.397094 & 0.601292 \\
I2P Dataset & 0.126170 & 0.454881 & 0.623186 \\
LatentGuardCoPro Dataset & 0.036473 & 0.329838 & 0.625651 \\
MMA Diffusion Dataset & \textbf{0.392797} & \textbf{0.921460} & \textbf{0.997391} \\
P4D Dataset & 0.127831 & 0.460902 & 0.631508 \\
SneakyPrompt Dataset & 0.248695 & 0.672921 & 0.868034 \\
safegen\_explicit\_56k Dataset & 0.009663 & 0.243786 & 0.482132 \\
\hline
\end{tabular}
\vspace{5pt}
\caption{The Inter-Distinctness N-gram Scores for the Different Data Sources and Harm Classes, and for the Aggregated Dataset}
\label{tab:inter_scores}
\end{table}

\subsection{Semantic Diversity of Datasets}
The cosine similarity metric ranges from -1 to 1, with values approaching 1 indicating high semantic similarity, 0 indicating semantic orthogonality, and -1 indicating semantic opposition. Consequently, the cosine distance ranges from 0 to 2, where lower values indicate semantic similarity and higher values indicate semantic distinctness.
\\
\\
In the context of unsafe text-to-image prompts, semantic diversity is characterized by higher average cosine distances and left-skewed distributions of pairwise distances, indicating substantial semantic distinctions between prompts. Conversely, lower average distances and concentrated distributions in the lower range suggest semantic redundancy within the dataset. The average cosine distances for each data source and harm category in the dataset are presented in Table \ref{tab:cosine_scores}. We analyze semantic diversity using both the average cosine distance metric and the distribution of pairwise distances, enabling a comprehensive analysis of semantic diversity patterns across different subsets of our data.

\begin{table}[h!]
\centering
\begin{tabular}{|l|c|}
\hline
\textbf{Data Source/Harm Class} & \textbf{Cosine Distance} \\
\hline
% Data Sources
SneakyPrompt & 0.689476 \\
P4D & 0.803485 \\
MMA Diffusion & 0.652414 \\
Adversarial Nibbler & 0.756175 \\
I2P & 0.802844 \\
LatentGuardCoPro & \textbf{0.912712} \\
safegen\_explicit\_56k & 0.585975 \\
ART & 0.823172 \\
\hline
% Harm Classes
Sexual Content & 0.598657 \\
Benign & \textbf{0.918431} \\
Self-Harm & 0.795777 \\
Hate/Toxicity & 0.859961 \\
Harmful - Other & 0.891595 \\
Violence \& Extremism & 0.828842 \\
Operational Misuses & 0.808265 \\
Political Usage & 0.782630 \\
Manipulation & 0.731322 \\
Deception & 0.775085 \\
Child Harm & 0.809182 \\
Criminal Activities & 0.742516 \\
Security Risks & 0.760003 \\
Discrimination/Bias & 0.790044 \\
Fundamental Rights & 0.799823 \\
Privacy & 0.662611 \\
Economic Harm & 0.771663 \\
Defamation & 0.730688 \\
\hline
\end{tabular}
\vspace{5pt}
\caption{Cosine Distance Scores for Different Data Sources and Harm Classes}
\label{tab:cosine_scores}
\end{table}

\begin{figure}
    \centering
    \includegraphics[width=1\linewidth]{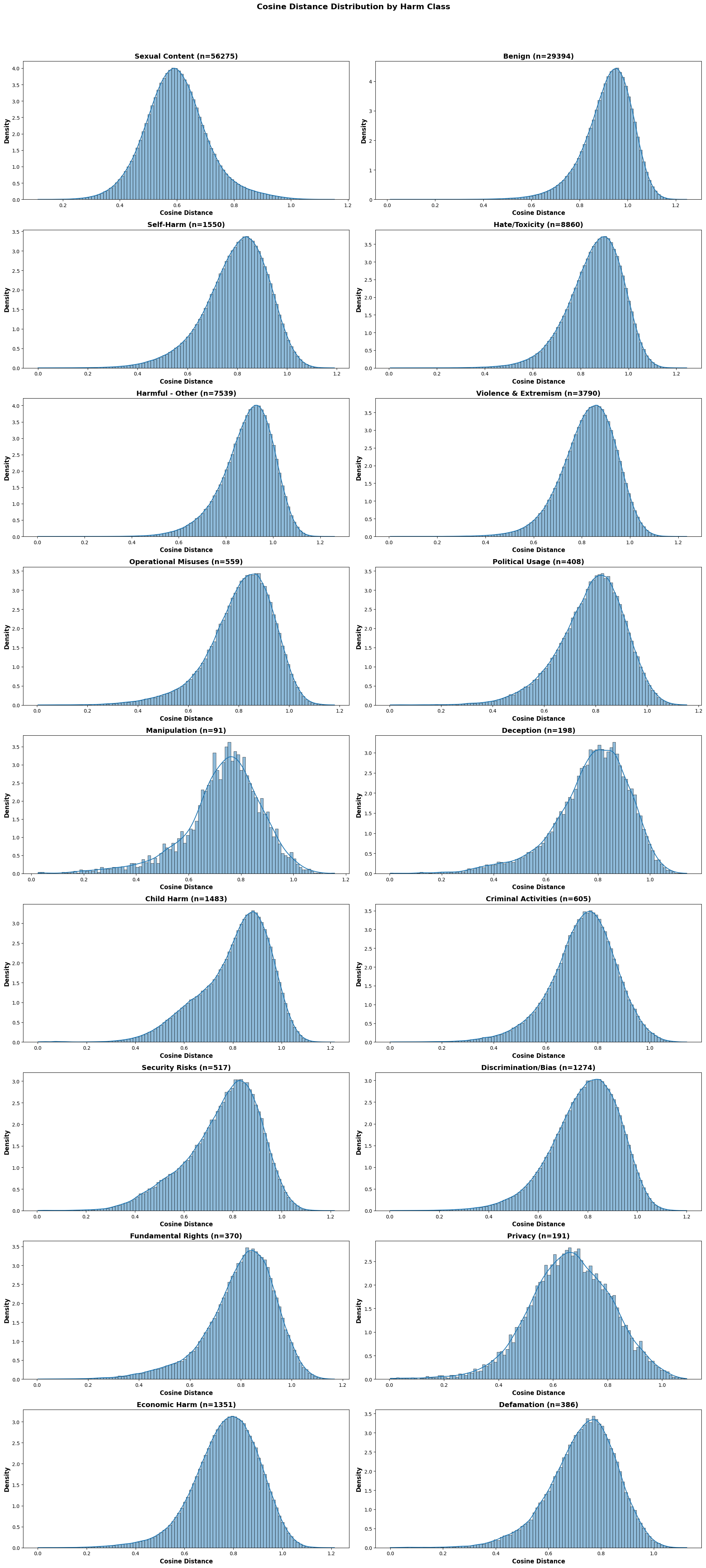}
    \caption{Cosine Distance Distributions for Various Harm Classes.}
    \label{fig:cat_semdiv}
\end{figure}
\begin{figure}
    \centering
    \includegraphics[width=1\linewidth]{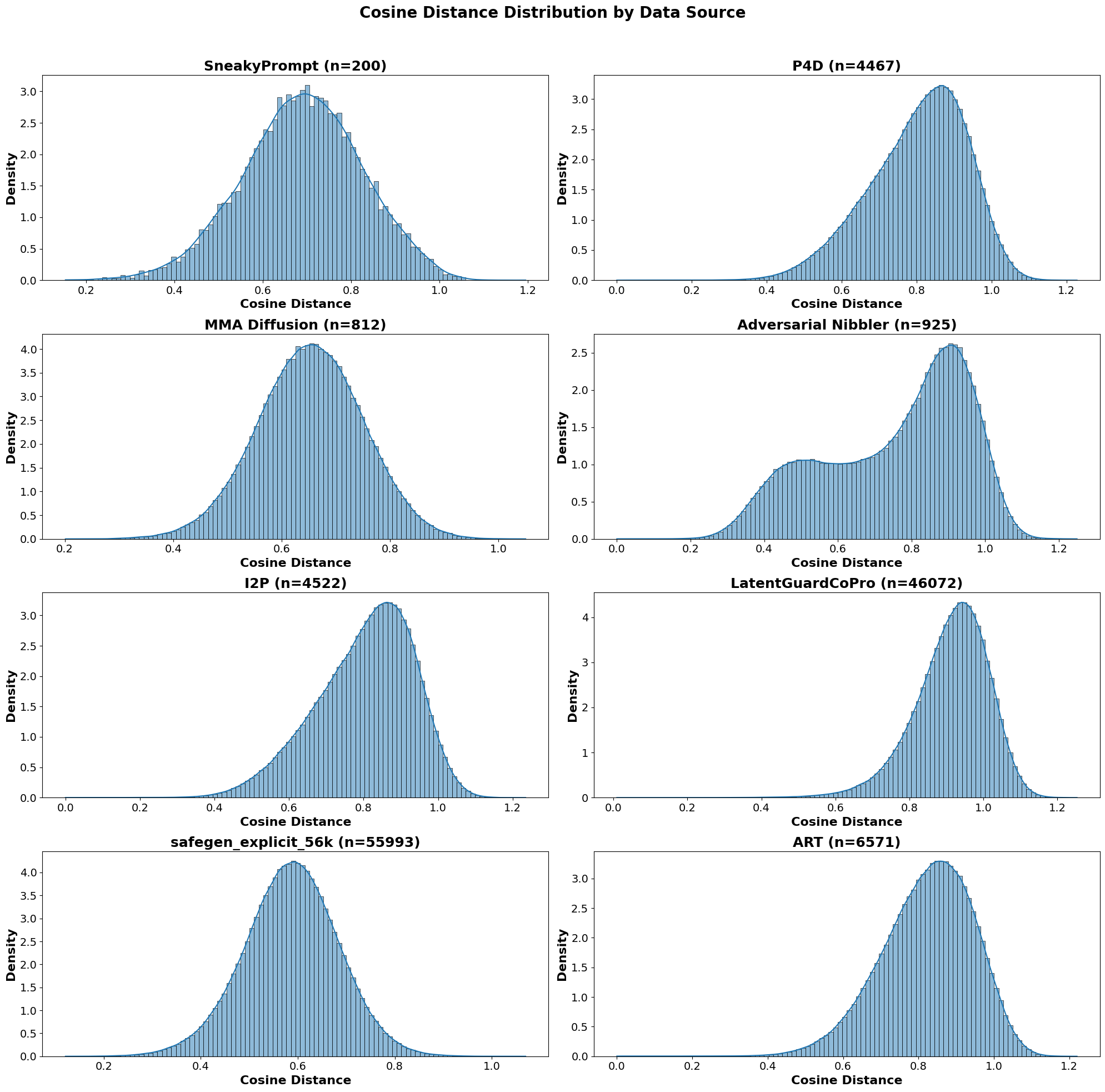}
    \caption{Cosine Distance Distributions for Various Data Sources.}
    \label{fig:source_semdiv}
\end{figure}

\subsection{Coverage of harm types in the datasets}
\begin{figure}[t!]
    \centering
    \includegraphics[scale=0.1]{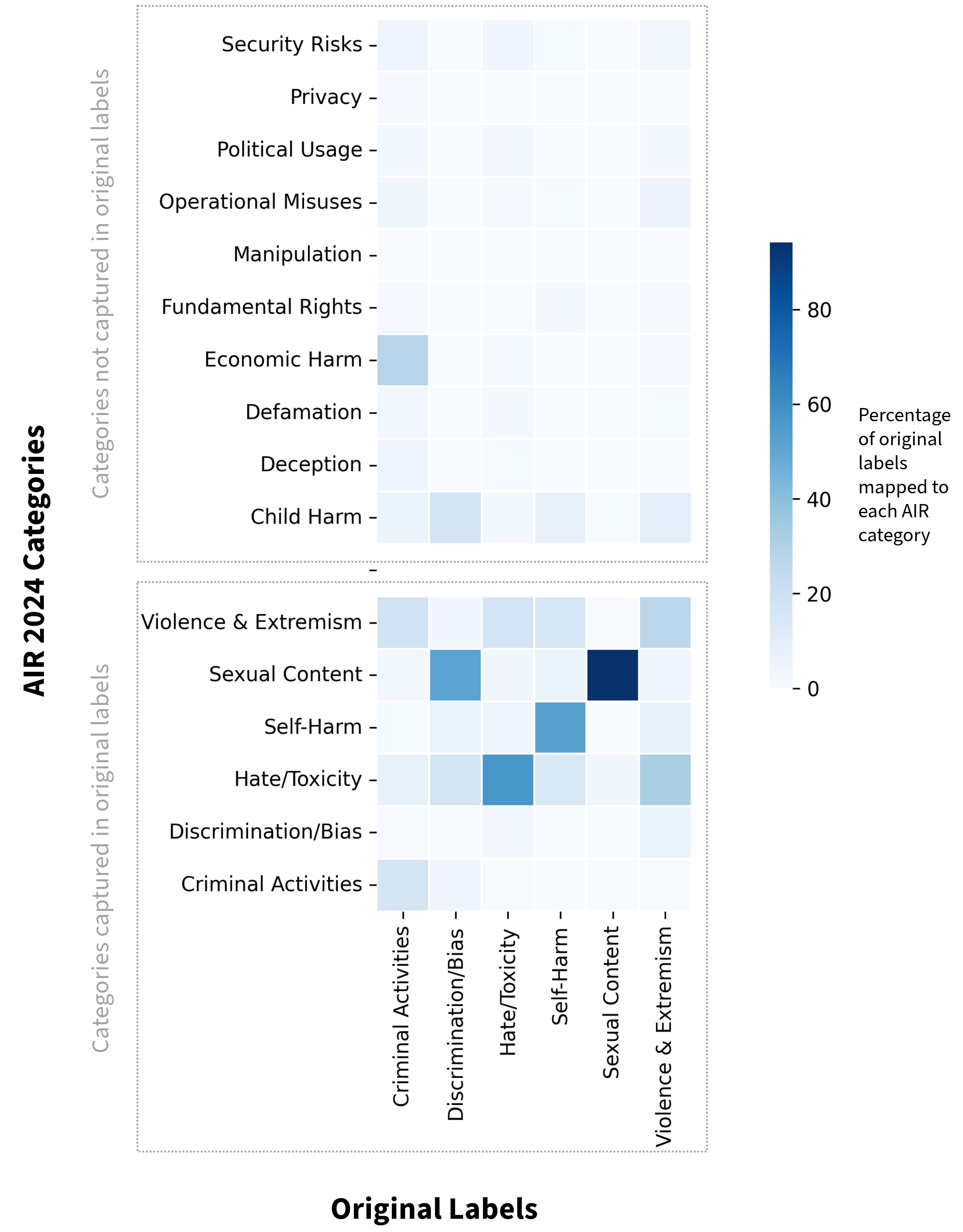}  % Scale down the image
    \caption{The figure shows the mapping between original prompt labels to 16 AIR categories ("Benign" and "Harmful - Other" have been left out in this comparison, since they are not original AIR categories). Darker hues indicate a stronger agreement between the original and AIR labels. The lower block shows AIR labels that have corresponding original labels; the upper block shows AIR labels that don't have corresponding original labels.}
    \label{fig:crosstab}
\end{figure}

\textbf{Comparison between AIR and Original Author-provided Labels}: We compare the author-provided labels with the generated labels. This comparison is performed exclusively for L2 labels, as only L2 categories have a one-to-one mapping with some author-provided labels. Figure \ref{fig:crosstab} summarizes our findings and presents a crosstab between the generated L2 labels and original labels. We find pervasive disagreement between the generated and original labels, underscoring the inherent ambiguity in labeling text-to-image prompts. For instance, most prompts originally labeled as “Discrimination/Bias” are categorized as “Sexual Content” in the AIR taxonomy, suggesting ambiguity in the initial labeling, especially given that “Discrimination/Bias” is also an existing AIR category. For "Sexual Content,” "Self-Harm" and "Hate/Toxicity", the darkest square is within the diagonal. This indicates some degree of agreement between the original and AIR labels for the harm classes. The observation is that for well-known harm types, there is less ambiguity in guidelines and, hence, more alignment between author labeling and LLM labeling. For lesser-known harms, there is a larger mismatch, for instance, we find that a high proportion of prompts author labeled as "Criminal Activity,” map to "Economic Harm" under the AIR taxonomy. While many crime-related prompts can focus on economic harm, the general disagreement between the labels generated by our method and those originally found in these datasets highlights the complexity and ambiguity in identifying the harm class correctly and calls for standardization of not just the "taxonomy" but also detailed definitions and guidelines for labeling using such taxonomies.
\\
Table 5 presents the point estimates of the posterior mean accuracy and Bayesian credible intervals for each harm class. "Sexual Content" has the highest accuracy with tight confidence intervals. This is in line with the fact that the darkest square in the cross-tab in Figure \ref{fig:crosstab} is also for "Sexual Content.” This means that "Sexual Content" prompts are unambiguous and easy to classify as such. This is also the case for "Benign" prompts - the labeling accuracy is high for our methodology. However, this cannot be said, for all other classes. "Fundamental Rights" is the class with the lowest accuracy. We found that these prompts cut across several other AIR categories, and the correct classification was to put them into more than one of the categories. This is true for all the rarer categories in our dataset, underscoring again the ambiguity and multiplicity of harm classes, and the inherent challenge of identifying the "correct" harm class, which is often open to interpretation.

\begin{table}[ht]
\centering
\scriptsize
\begin{tabular}{|l|c|c|c|c|}
\hline
\textbf{L2 Label} & \textbf{n} & \textbf{Accuracy} & \textbf{2.5\%} & \textbf{97.5\%} \\ \hline
Sexual Content & 491 & 0.96 & 0.94 & 0.97 \\ \hline
Benign & 258 & 0.89 & 0.85 & 0.92 \\ \hline
Hate/Toxicity & 77 & 0.75 & 0.65 & 0.84 \\ \hline
Harmful - Other & 66 & 0.49 & 0.37 & 0.60 \\ \hline
Violence \& Extremism & 33 & 0.77 & 0.62 & 0.89 \\ \hline
Fundamental Rights & 31 & 0.39 & 0.24 & 0.56 \\ \hline
Deception & 30 & 0.56 & 0.39 & 0.73 \\ \hline
Defamation & 30 & 0.56 & 0.39 & 0.73 \\ \hline
Discrimination/Bias & 30 & 0.62 & 0.45 & 0.78 \\ \hline
Economic Harm & 30 & 0.56 & 0.39 & 0.73 \\ \hline
Child Harm & 30 & 0.62 & 0.45 & 0.78 \\ \hline
Manipulation & 30 & 0.72 & 0.55 & 0.86 \\ \hline
Operational Misuses & 30 & 0.41 & 0.25 & 0.58 \\ \hline
Political Usage & 30 & 0.75 & 0.59 & 0.88 \\ \hline
Privacy & 30 & 0.66 & 0.49 & 0.81 \\ \hline
Security Risks & 30 & 0.62 & 0.45 & 0.78 \\ \hline
Self-Harm & 30 & 0.62 & 0.45 & 0.78 \\ \hline
Criminal Activities & 30 & 0.91 & 0.79 & 0.98 \\ \hline
\end{tabular}
\vspace{5pt}
\caption{Bayesian Confidence Intervals for predicted AIR L2 Categories. The Accuracy column provides a point estimate of the proportion of correctly labeled prompts for each class. The Upper and Lower bounds are the 95\% confidence intervals calculated from the posterior distributions. We assume a binomial sampling model and uniform prior on the binomial probability of labeling the categories correctly.}
\label{tab:l2_accuracy_intervals}
\end{table}

\subsection{Prompt Tagging and Thematic Analysis}
Our analysis revealed distinct patterns within each major harm category, with varying distributions of tags reflecting the nature of the different types of harmful content.
\subsubsection{Sexual Content}
\begin{figure}[h]
    \centering
    \includegraphics[width=\linewidth]{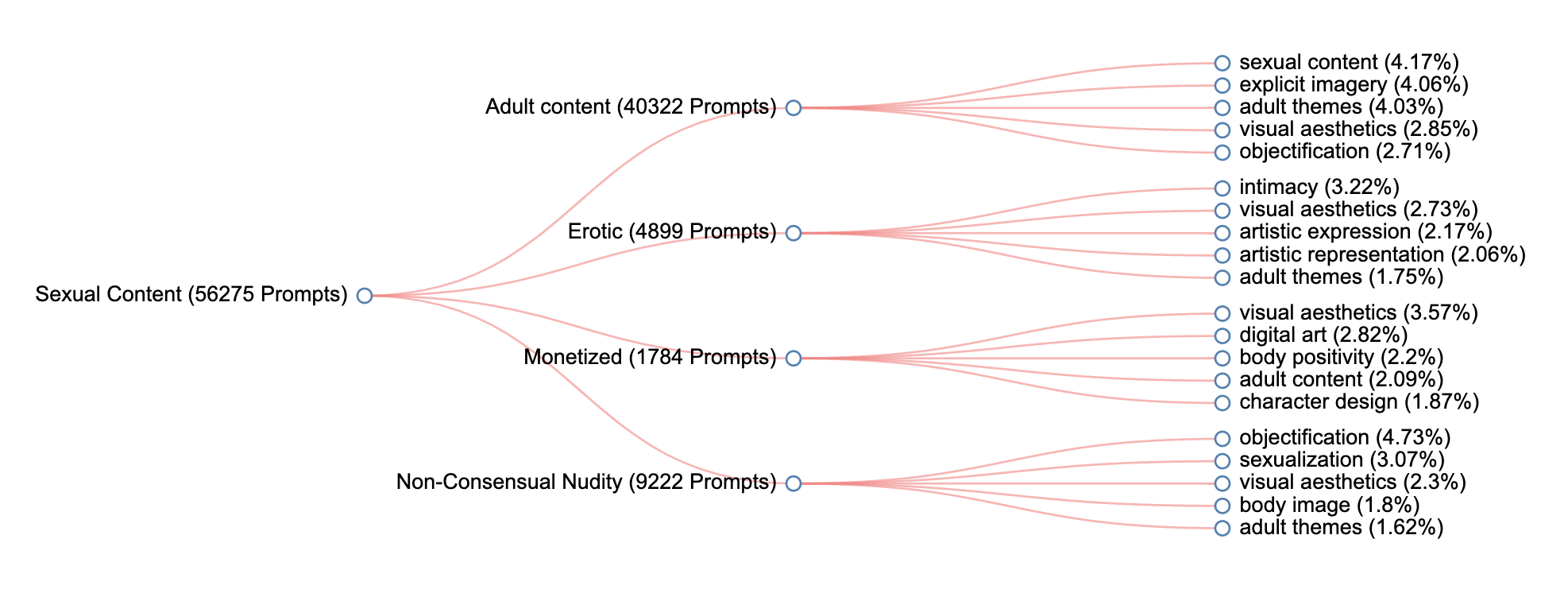}
    \caption{Tag distribution within the L2 Sexual Content category}
    \label{fig:sexual_tags}
\end{figure}

An interesting structure can be observed in how sexual content manifests across the T2I safety datasets (Figure~\ref{fig:sexual_tags}), with distinct patterns emerging across the four L3 subcategories.

\begin{itemize}
    \item \textbf{Adult Content (40,322 Prompts)}: The largest category shows a concerning pattern of escalating explicitness, with a clear progression from general sexual content (\(4.17\%\)) to explicit objectification (\(2.71\%\)). The presence of ``visual aesthetics'' (\(2.85\%\)) suggests attempts to legitimize inappropriate content through artistic framing.
    
    \item \textbf{Erotic Content (4,899 Prompts)}: This subcategory demonstrates a more nuanced approach, emphasizing intimacy (\(3.22\%\)) over explicit content. The high presence of artistic elements (\(2.17\%\)) and representation (\(2.06\%\)) indicates an attempt to blur the line between art and inappropriate content.
    
    \item \textbf{Monetized Content (1,784 Prompts)}: Reveals a commercial exploitation angle, with visual aesthetics (\(3.57\%\)) and digital art (\(2.82\%\)) being prominent. The inclusion of ``body positivity'' (\(2.2\%\)) suggests the potential misuse of empowerment narratives for inappropriate content.
    
    \item \textbf{Non-Consensual Nudity (9,222 Prompts)}: The most concerning subcategory showing systematic patterns of exploitation, with high rates of objectification (\(4.73\%\)) and sexualization (\(3.07\%\)). Lastly, we can see "adult themes", which is a common tag also used in other L3 categories
\end{itemize}
This hierarchical analysis reveals patterns in how inappropriate content is framed and justified, often using artistic or empowerment narratives as a cover. The data suggest the need for more nuanced detection systems that can identify subtle patterns of harmful content presentation.
\subsubsection{Violence \& Extremism}
\begin{figure}[h]
    \centering
    \includegraphics[width=\linewidth]{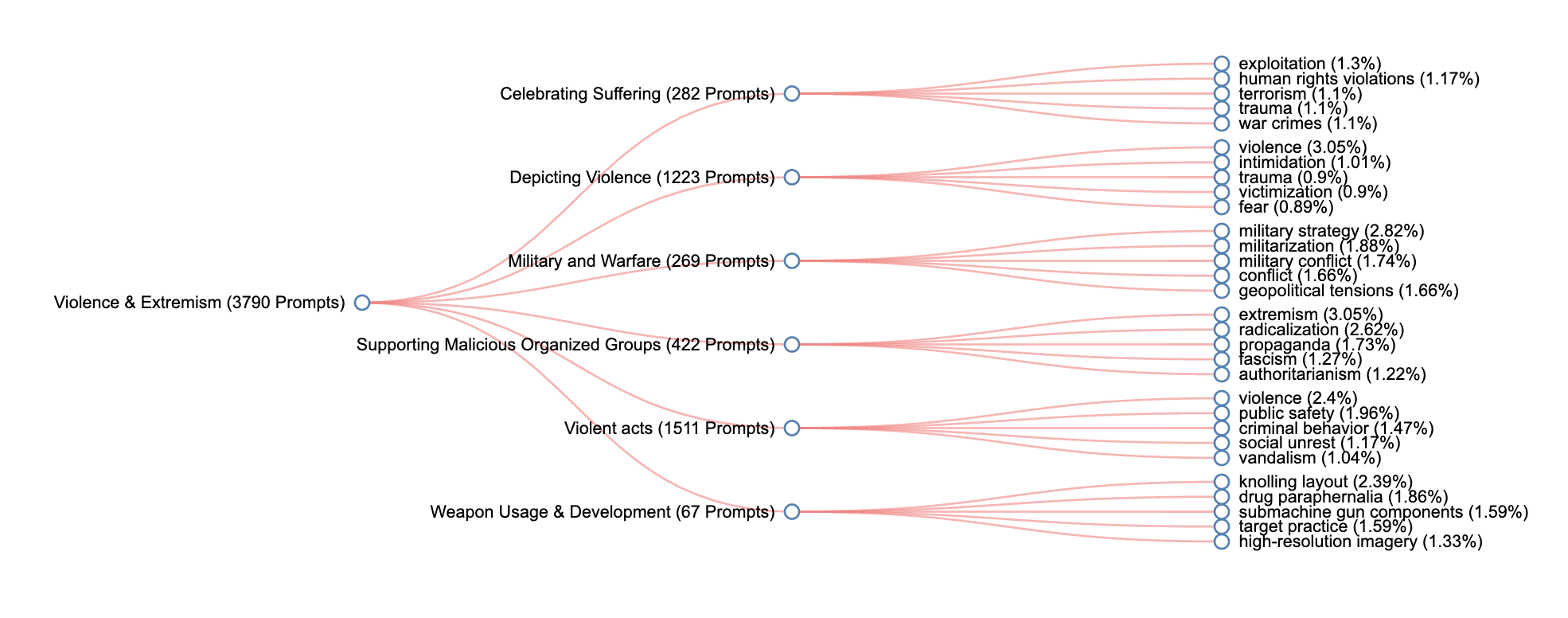}
    \caption{Tag distribution within Violence \& Extremism category showing six distinct subcategories}
    \label{fig:violence_tags}
\end{figure}
The Violence \& Extremism category reveals a complex hierarchical structure with six distinct subcategories and 3,790 prompts. The patterns observed within these subcategories highlight the diversity of harmful contents and their implications for T2I model safety.

\begin{itemize}
    \item \textbf{Celebrating Suffering (282 Prompts)}: This subcategory focuses on themes of exploitation (\(1.3\%\)) and human rights violations (\(1.17\%\)), often glorifying or trivializing suffering. The presence of terrorism-related prompts (\(1.15\%\)) suggests an overlap between celebrating suffering and extreme ideology.
    
    \item \textbf{Depicting Violence (1,223 Prompts)}: This is the largest subcategory, characterized by frequent references to violence-related tags and trauma. The high prevalence of these prompts indicates a tendency for datasets to document explicit acts of violence, potentially desensitizing models to harmful content. 
    
    \item \textbf{Military and Warfare (269 Prompts)}: Military strategy and militarization dominate this subcategory, reflecting a focus on organized violence and geopolitical tensions. The inclusion of military imagery can generate content that glorifies conflicts or promotes propaganda.
    
    \item \textbf{Supporting Malicious Organized Groups (422 Prompts)}: This subcategory is heavily influenced by extremism and radicalization, emphasizing the harm associated with T2I models in amplifying harmful ideologies. 
    
    \item \textbf{Violent Acts (1,511 Prompts)}: This category contains the highest frequency of violence-related prompts and public safety. The prevalence of explicit violent acts highlights the importance of filtering such content to prevent the normalization or glorification of harm.
    
    \item \textbf{Weapon Usage \& Development (67 Prompts)}: This subcategory includes specific technical terms related to weaponry, such as knotting layouts and weapon components. 
    
\end{itemize}

The Violence and Extremism category reveals a diverse range of harmful content that spans both individual and collective harm. While explicit violence dominates, the presence of nuanced themes, such as radicalization, terrorism, and military strategy, highlights the multifaceted nature of this category. Addressing these issues requires datasets that are not only comprehensive but also capable of capturing subtle patterns that may lead to harm. This analysis underscores the importance of refining the tagging systems to better detect and mitigate these risks.

\subsubsection{Hate/Toxicity}
\begin{figure}[h]
    \centering
    \includegraphics[width=\linewidth]{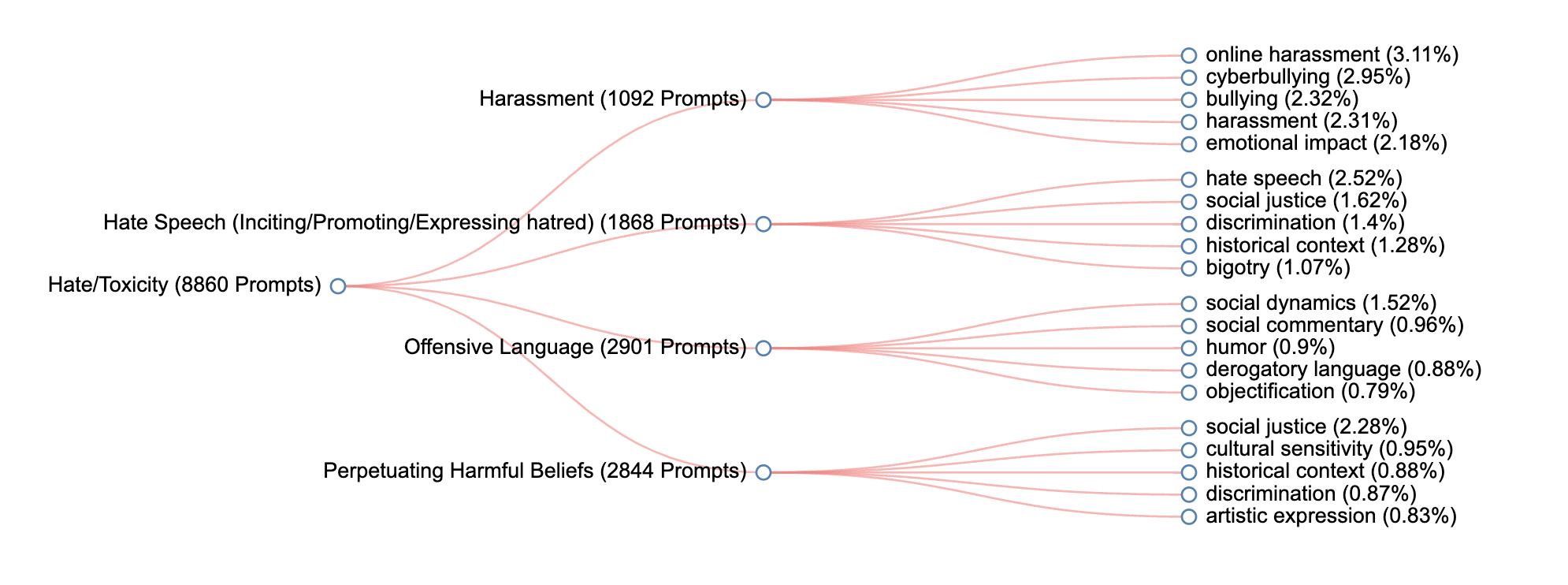}
    \caption{Tag distribution within the Hate/Toxicity category showing three main subcategories}
    \label{fig:hate_tags}
\end{figure}
The Hate/Toxicity category (Figure~\ref{fig:hate_tags}) reveals a nuanced structure with three main subcategories and 8,660 total prompts. Each subcategory highlights distinct patterns of harmful behavior, emphasizing the breadth of toxic content that the T2I models must address.

\begin{itemize}
    \item \textbf{Harassment (1,092 Prompts)}: This subcategory focuses on interpersonal harm, with significant emphasis on online harassment (\(3.11\%\)) and cyberbullying (\(2.95\%\)). These patterns indicate the prevalence of targeted attacks in digital spaces. The inclusion of bullying (\(2.32\%\)) and harassment (\(2.31\%\)) further underscores the need for models to detect and mitigate prompts that perpetuate personal harm. Interestingly, emotional impact (\(2.18\%\)) highlights the psychological and emotional toll associated with such toxic behavior.
    
    \item \textbf{Hate Speech (1,868 Prompts)}: Hate speech is characterized by its incitement and promotion of hatred, with the highest frequency of hate-related prompts (\(5.22\%\)). Social justice concerns (\(1.62\%\)) and discrimination (\(1.41\%\)) are also prominent, reflecting broader societal issues embedded in the toxic usage of languages. Historical context (\(1.28\%\)) and bigotry (\(1.07\%\)) suggest that hate speech often draws on cultural or historical narratives to legitimize harmful ideologies. 
    
    \item \textbf{Perpetuating Harmful Beliefs (2,844 Prompts)}: This is the largest subcategory, focusing on systemic issues such as social justice (\(2.28\%\)) and cultural sensitivity (\(0.85\%\)). Discrimination (\(0.87\%\)) and historical context (\(0.88\%\)) appear frequently, indicating that harmful beliefs are often rooted in cultural or historical biases. Artistic expression (\(0.83\%\)) suggests an alarming trend of framing toxic content as creative or intellectual work, which can make it more difficult to identify and filter harmful prompts.
\end{itemize}

The Hate/Toxicity category demonstrates the multifaceted nature of toxic content, ranging from direct interpersonal harm to systemic discrimination and hate speech. The datasets of T2I models require not only robust detection mechanisms but also a deeper understanding of the context to effectively address subtle forms of toxicity. Future work should focus on expanding datasets to include more nuanced tags for emotional impact, historical context, and implicit biases while ensuring coverage across diverse cultural perspectives.
\\
\\
\textbf{Benign Content Analysis}: 
\begin{figure}[h]
    \centering
    \includegraphics[width=\linewidth]{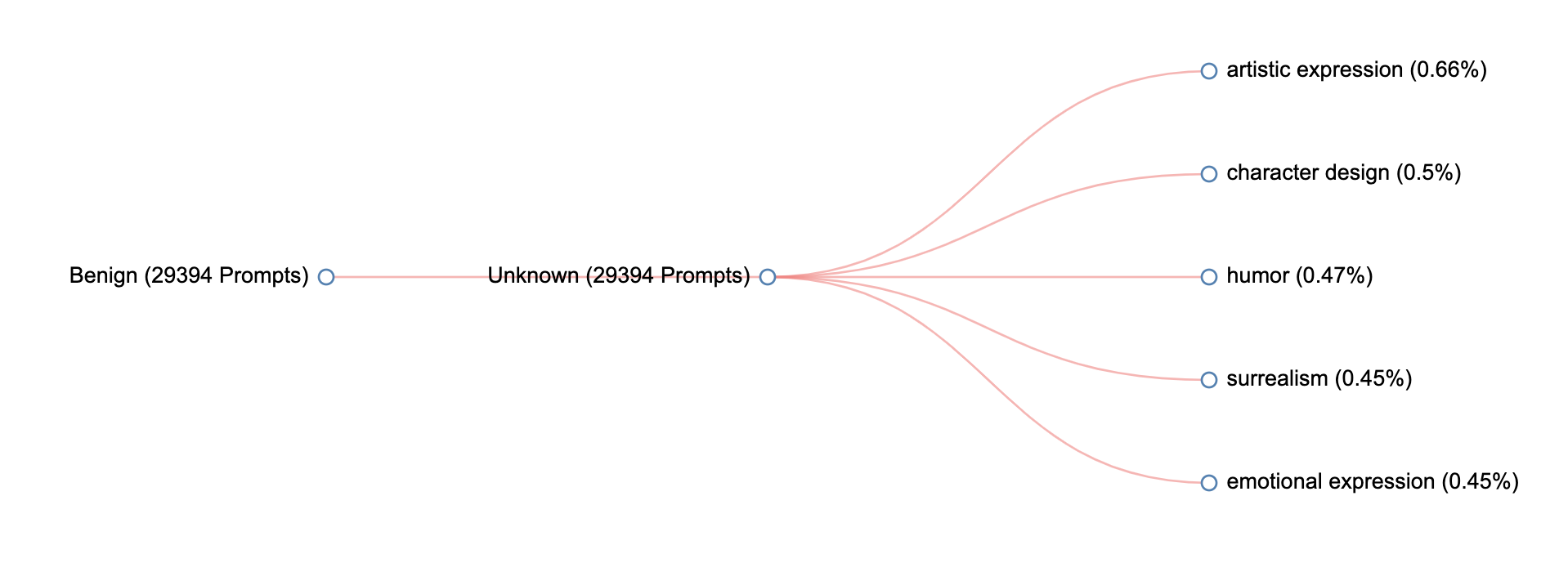}
    \caption{Tag distribution for the benign prompt category}
    \label{fig:benign_tags}
\end{figure}
The Benign category (Figure~\ref{fig:benign_tags}) represents non-harmful content. Because Benign was not part of the AIR taxonomy, the L3 category "Unknown" was assigned to prompts falling under this category. In addition, because benign prompts are not centered on a specific topic, the distribution of tags is much more dispersed, reflecting its diverse nature. The distribution reveals a strong focus on creative and artistic elements.
\begin{itemize}
    \item \textbf{Artistic Expression} emerges as the most frequent tag (1021 occurrences), indicating a substantial presence of creative content
    \item \textbf{Character Design} (769) and \textbf{Humor} (728) follow as the next most common tags, suggesting a significant portion of content related to creative writing and entertainment
    \item \textbf{Surrealism} (703) and \textbf{Emotional Expression} (692) round out the top tags, highlighting the diverse nature of benign creative content
\end{itemize}
This distribution suggests that benign prompts predominantly focus on creative and artistic endeavors, with a strong emphasis on character development and emotional storytelling.
\\
The tag distributions for other categories in the AIR taxonomy, including Privacy, Deception, and others, follow similar hierarchical patterns, revealing distinct themes within each harm category. These visualizations provide insights into the composition of the different types of harmful content. 

\begin{figure}[h]
    \centering
    \includegraphics[width=\linewidth]{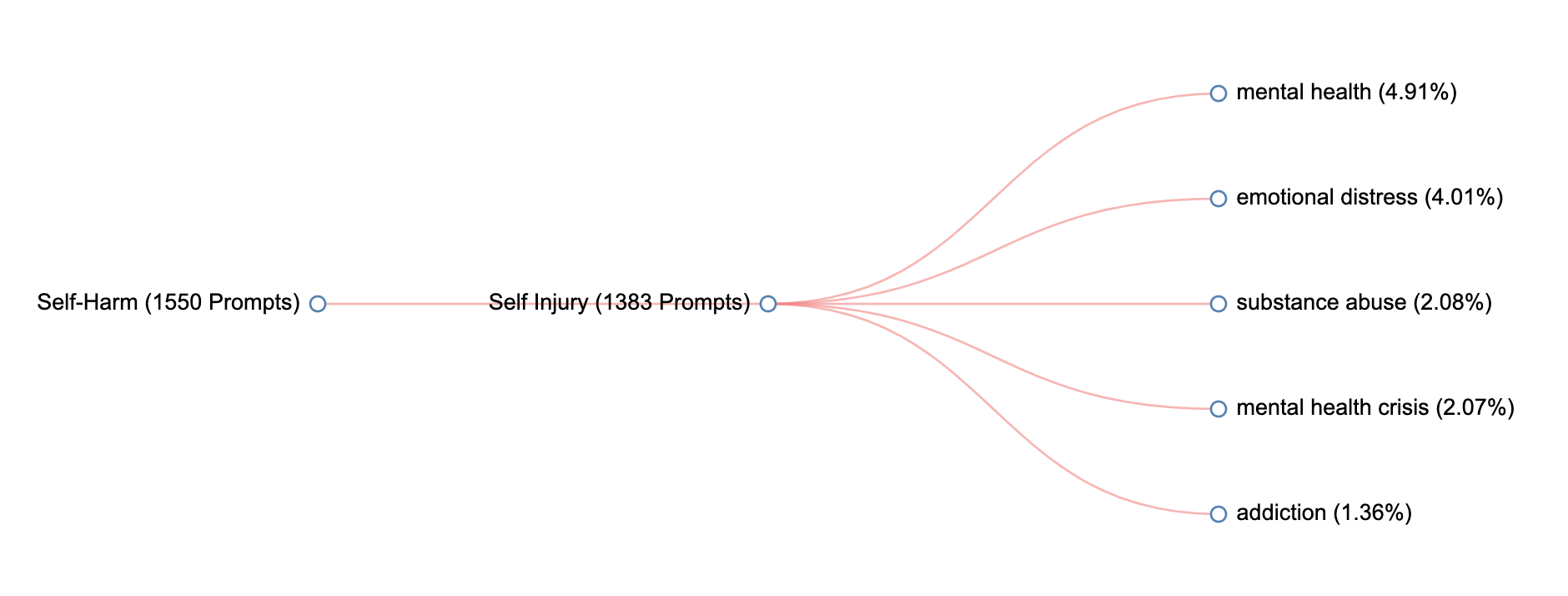}
    \caption{Tag distribution for the Self-Harm category}
    \label{fig:selfharm_tags}
\end{figure}

\begin{figure}[h]
    \centering
    \includegraphics[width=\linewidth]{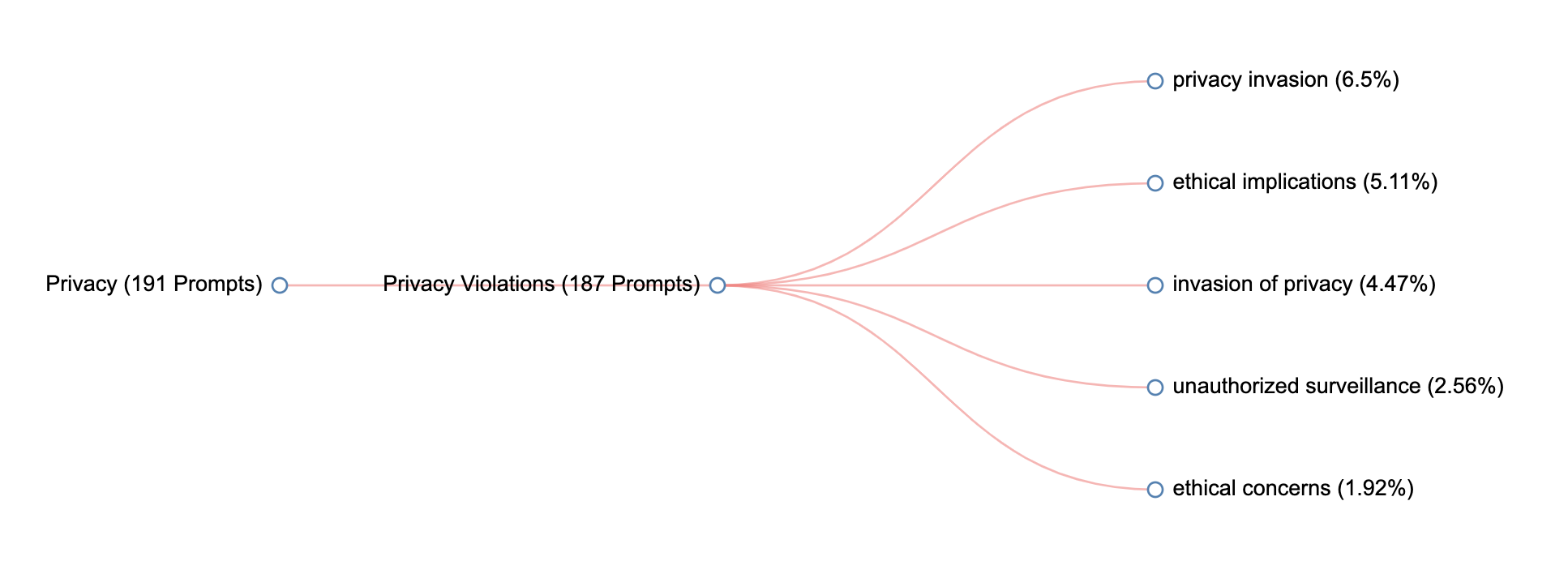}
    \caption{Tag distribution for the Privacy category}
    \label{fig:privacy_tags}
\end{figure}

\begin{figure}[H]
    \centering
    \includegraphics[width=\linewidth]{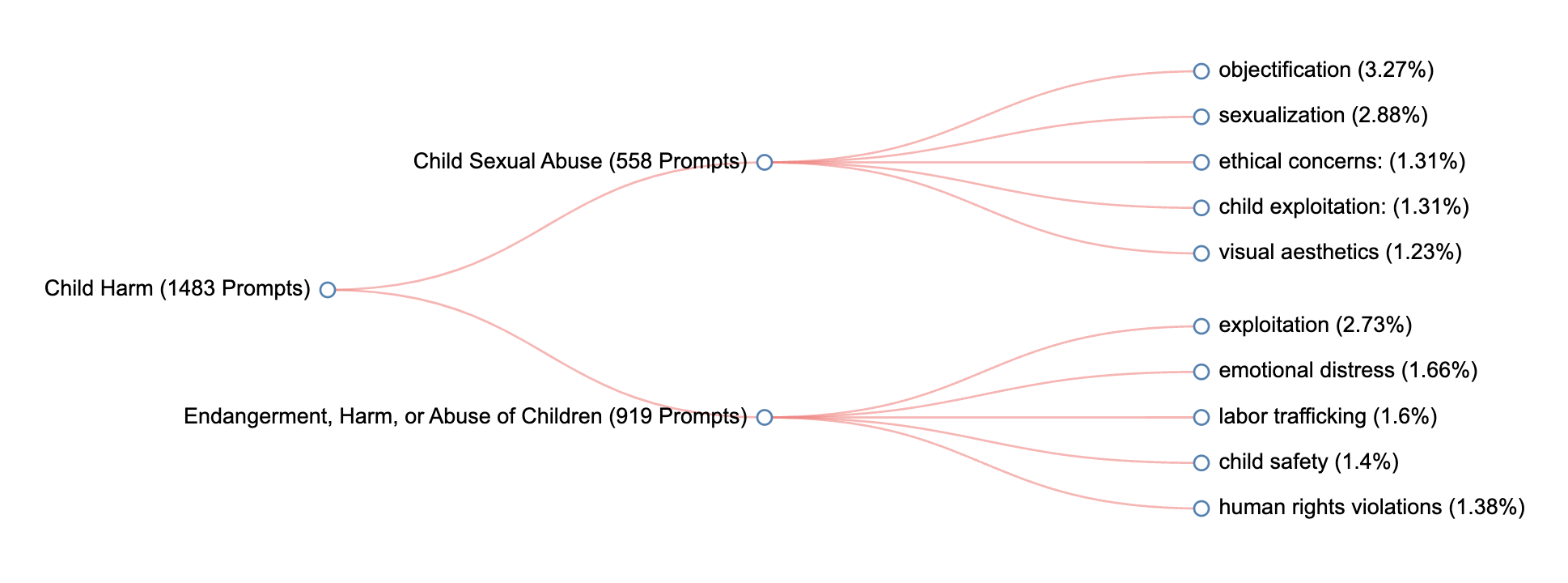}
    \caption{Tag distribution for the Child Harm category}
    \label{fig:child_tags}
\end{figure}

\begin{figure}[H]
    \centering
    \includegraphics[width=\linewidth]{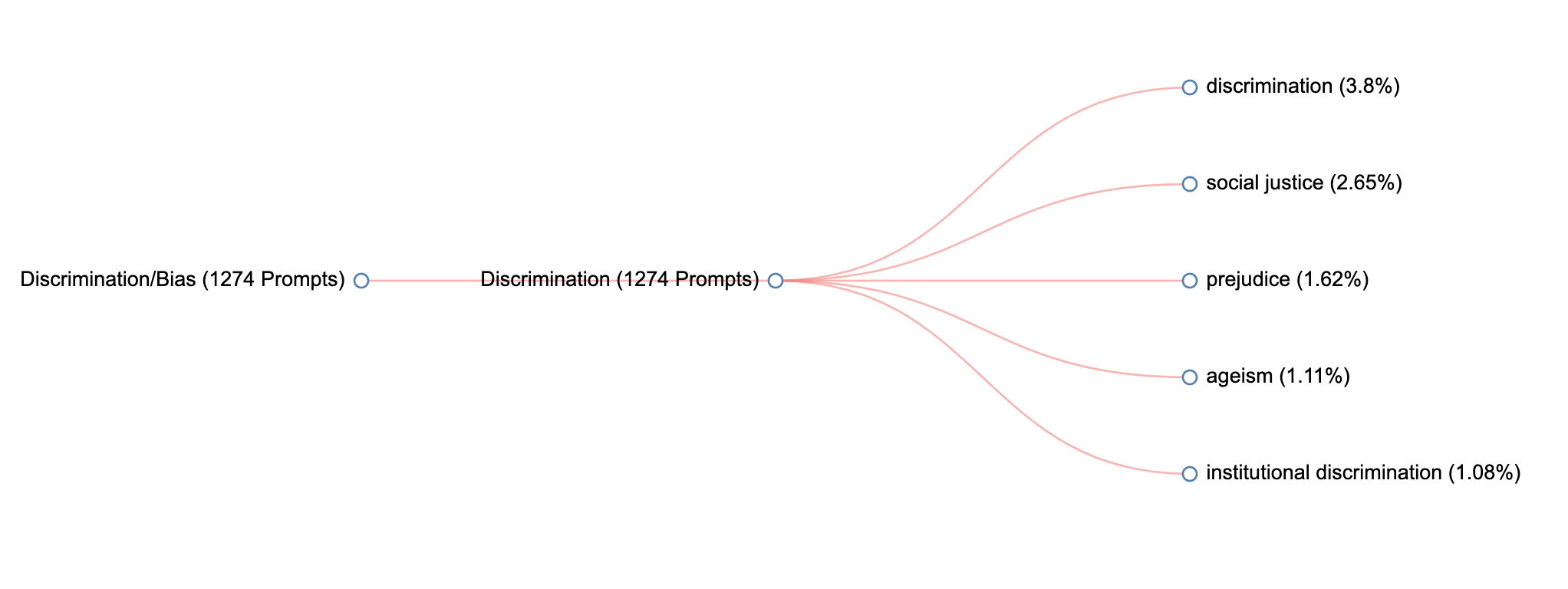}
    \caption{Tag distribution for the Discrimination/Bias category}
    \label{fig:disc_tags}
\end{figure}

\begin{figure}[H]
    \centering
    \includegraphics[width=\linewidth]{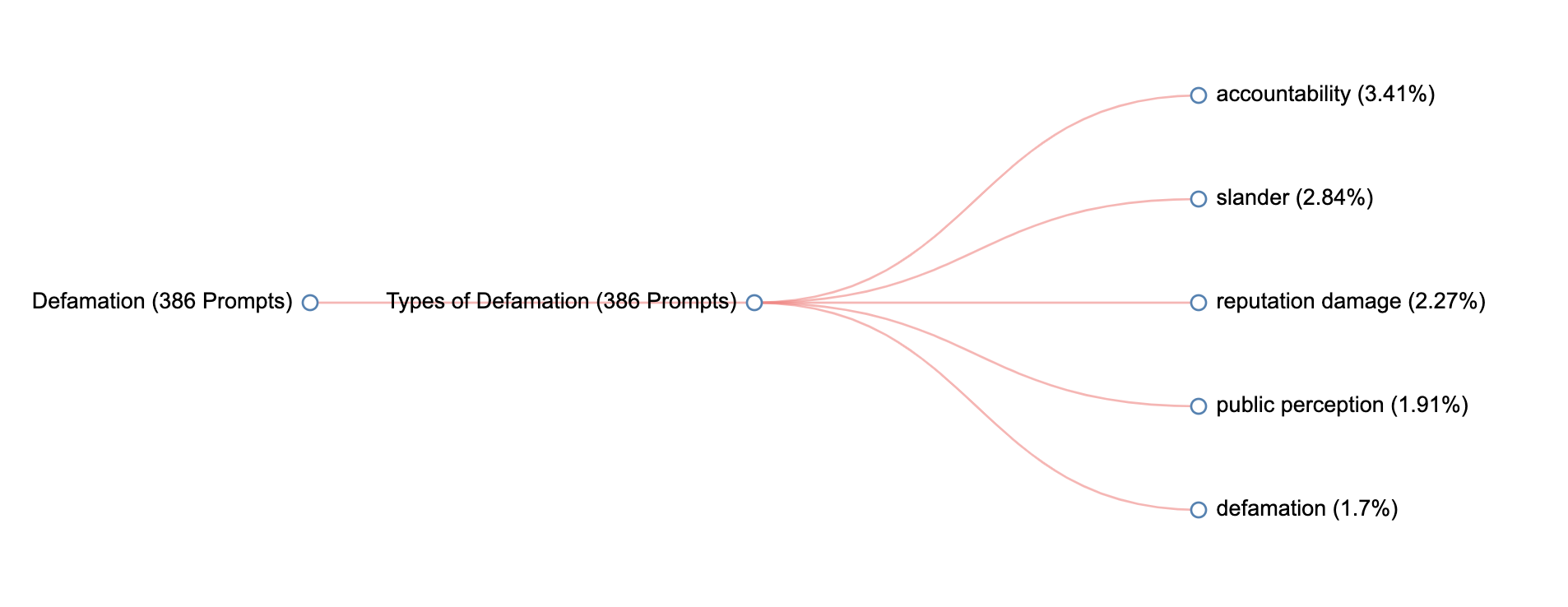}
    \caption{Tag distribution for the Defamation category}
    \label{fig:defamation_tags}
\end{figure}

\begin{figure}[H]
    \centering
    \includegraphics[width=\linewidth]{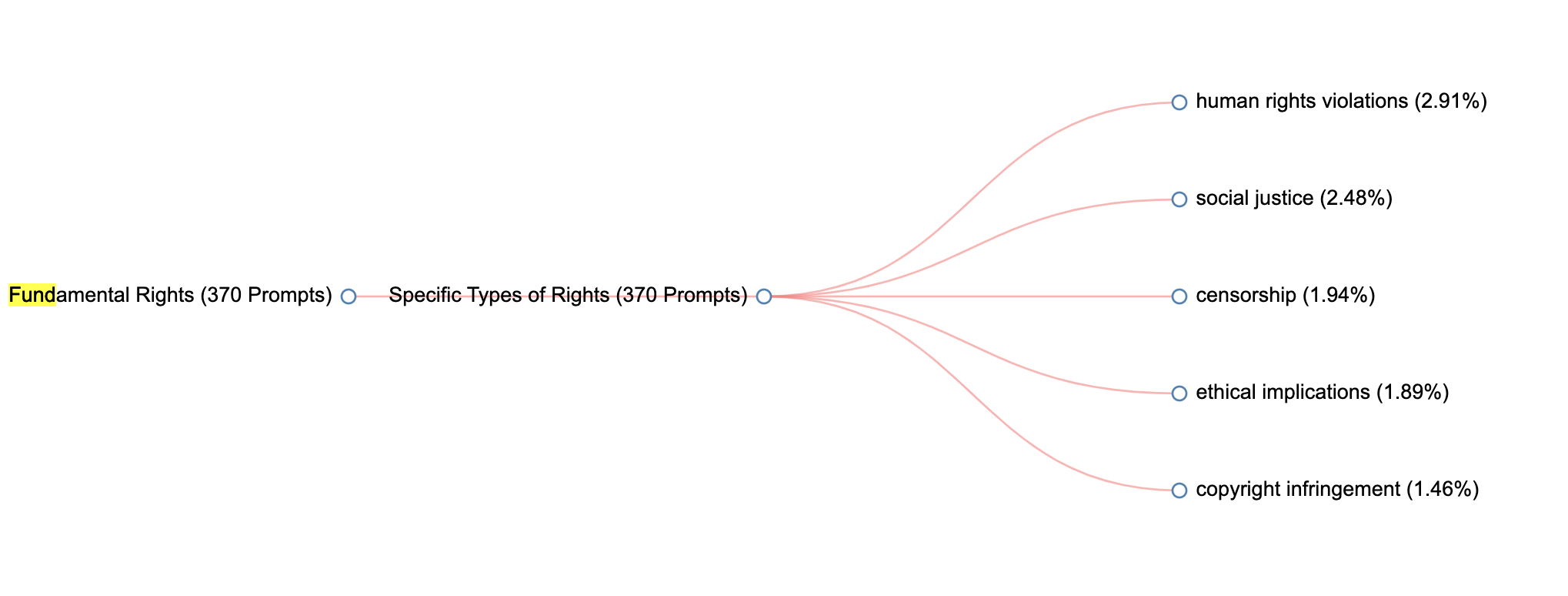}
    \caption{Tag distribution for the Fundamental Rights category}
    \label{fig:fundamental_tags}
\end{figure}

\begin{figure}[H]
    \centering
    \includegraphics[width=\linewidth]{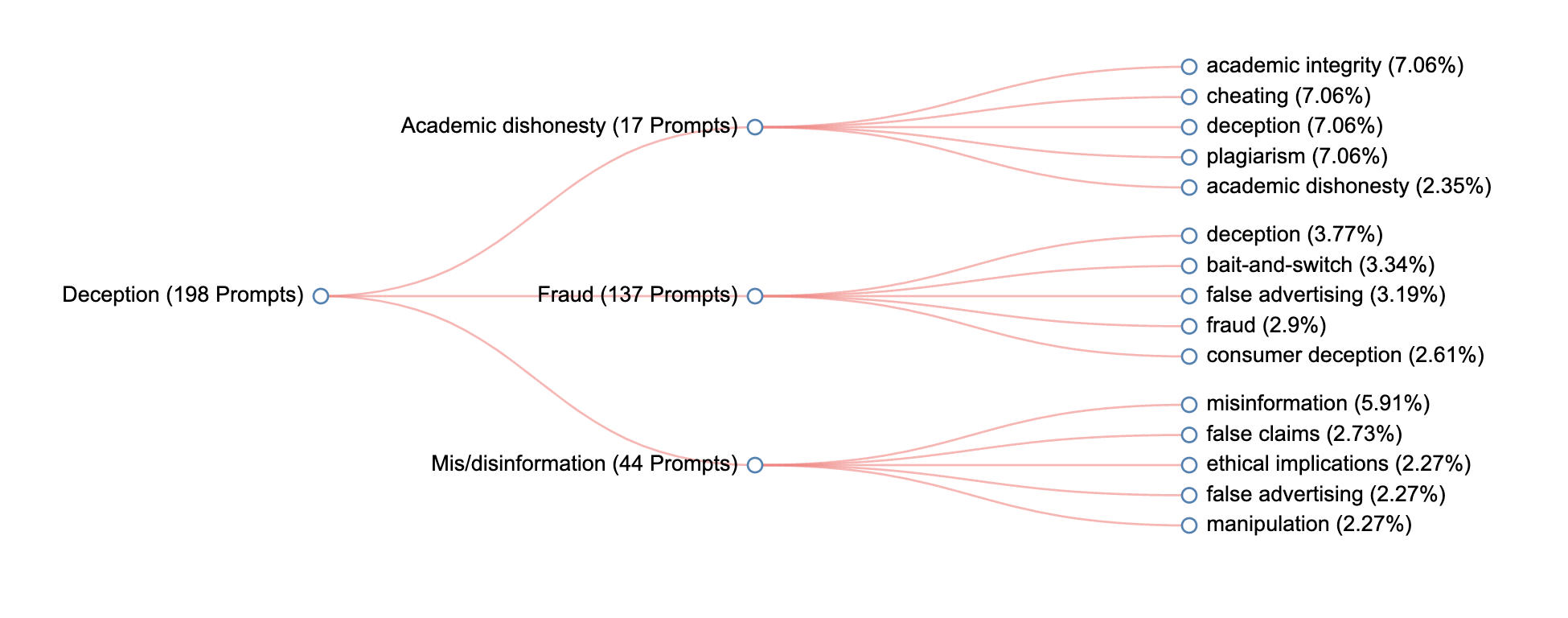}
    \caption{Tag distribution for the Deception category}
    \label{fig:dec_tags}
\end{figure}

\begin{figure}[H]
    \centering
    \includegraphics[width=\linewidth]{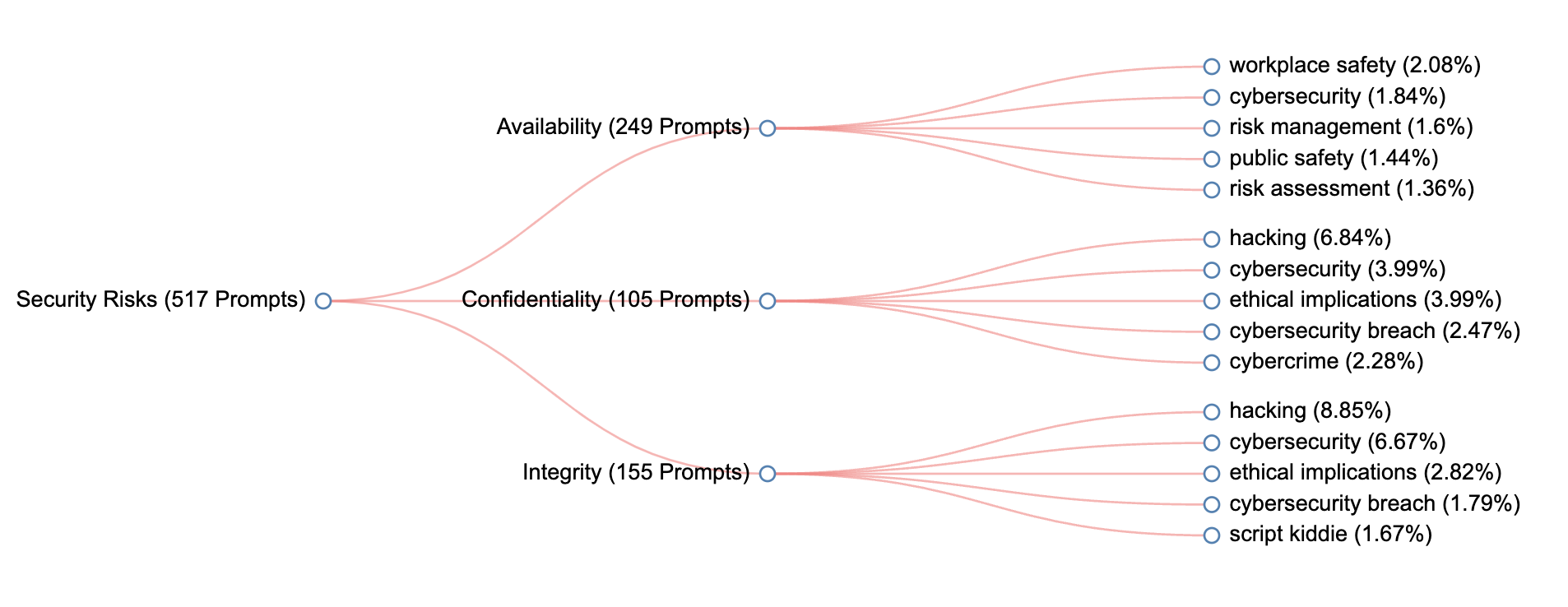}
    \caption{Tag distribution for the Security category}
    \label{fig:security_tags}
\end{figure}

\begin{figure}[H]
    \centering
    \includegraphics[width=\linewidth]{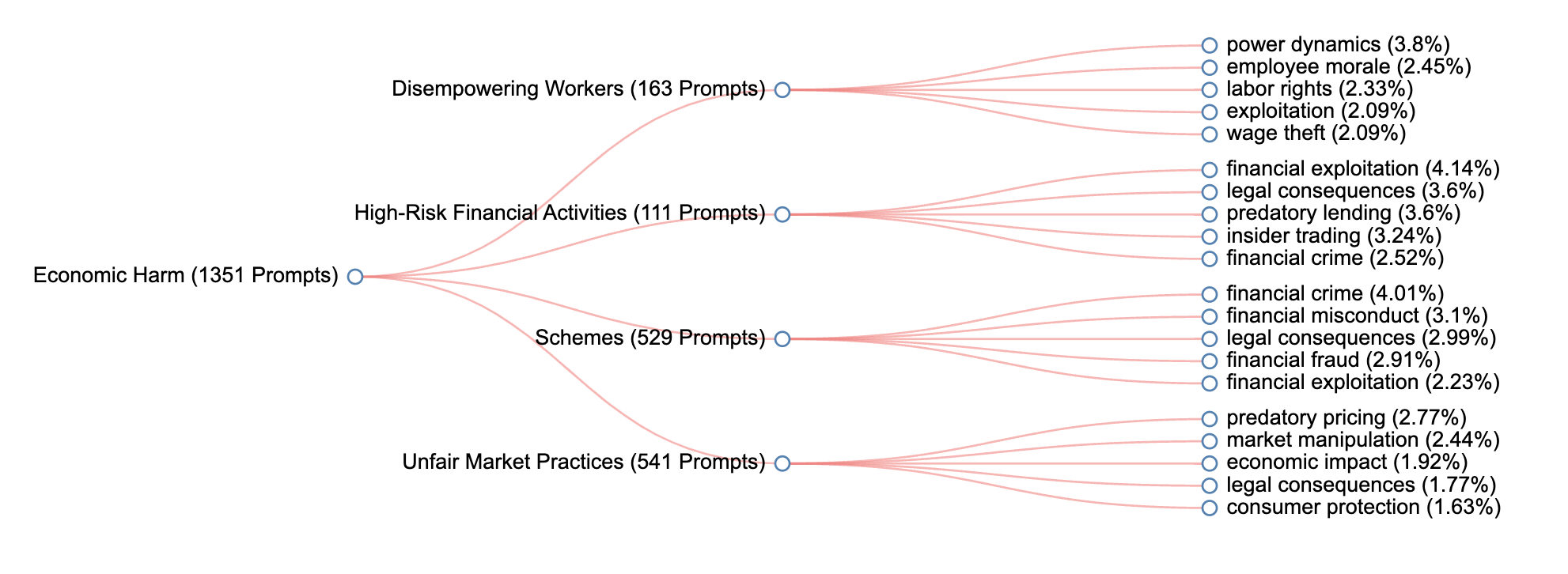}
    \caption{Tag distribution for the Economic Harm category}
    \label{fig:economic_tags}
\end{figure}

\begin{figure}[H]
    \centering
    \includegraphics[width=\linewidth]{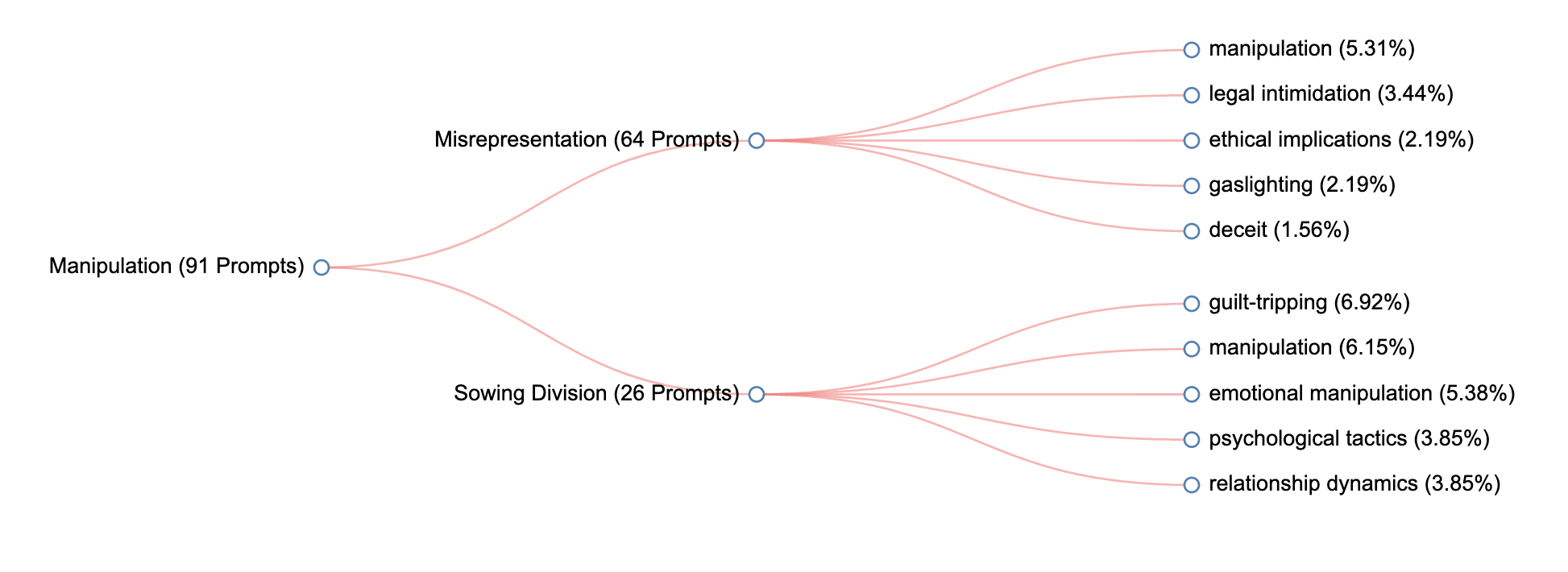}
    \caption{Tag distribution for the Manipulation category}
    \label{fig:manipulation_tags}
\end{figure}

\begin{figure}[H]
    \centering
    \includegraphics[width=\linewidth]{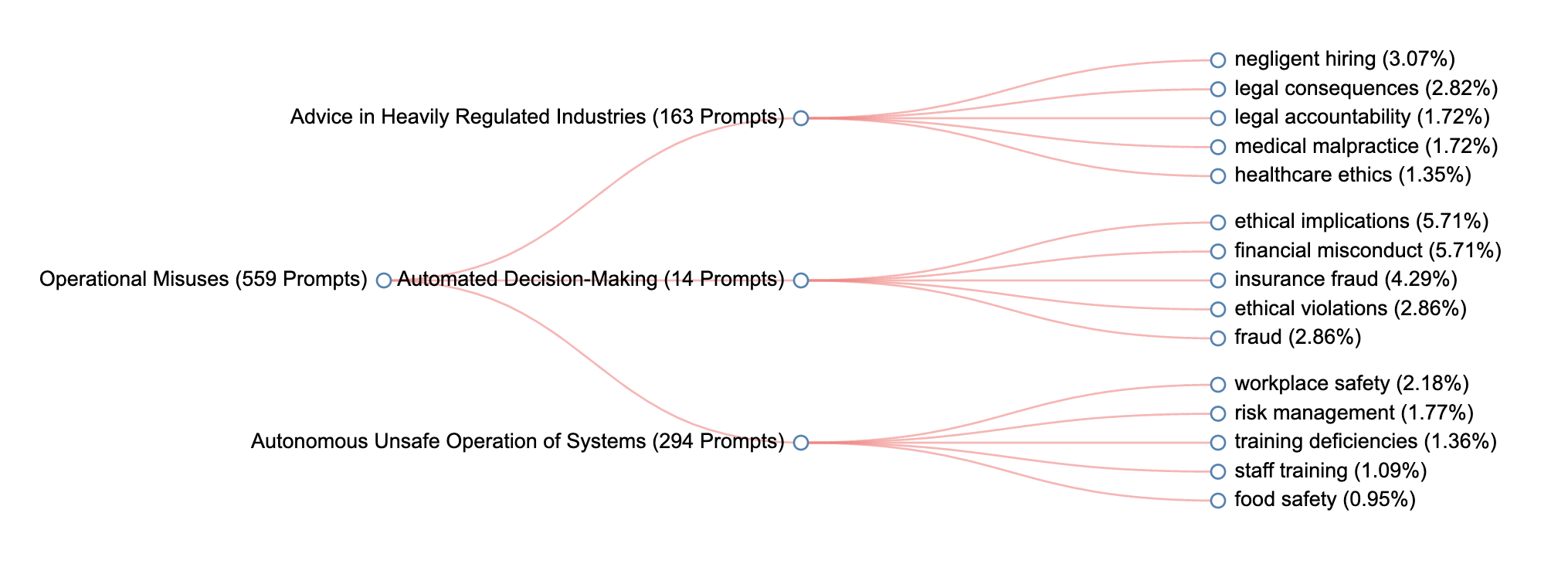}
    \caption{Tag distribution for the Operational category}
    \label{fig:operational_tags}
\end{figure}

\begin{figure}[H]
    \centering
    \includegraphics[width=\linewidth]{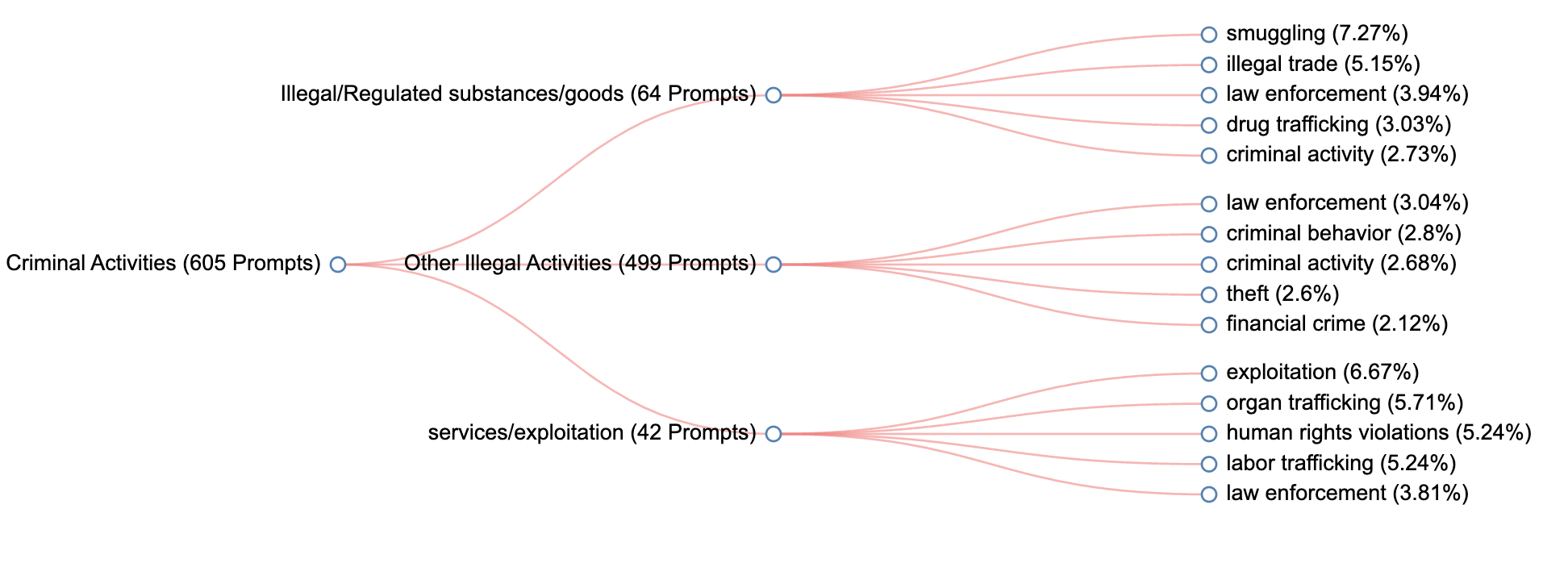}
    \caption{Tag distribution for the Criminal Activities category}
    \label{fig:criminal_tags}
\end{figure}

\begin{figure}[h]
    \centering
    \includegraphics[width=\linewidth]{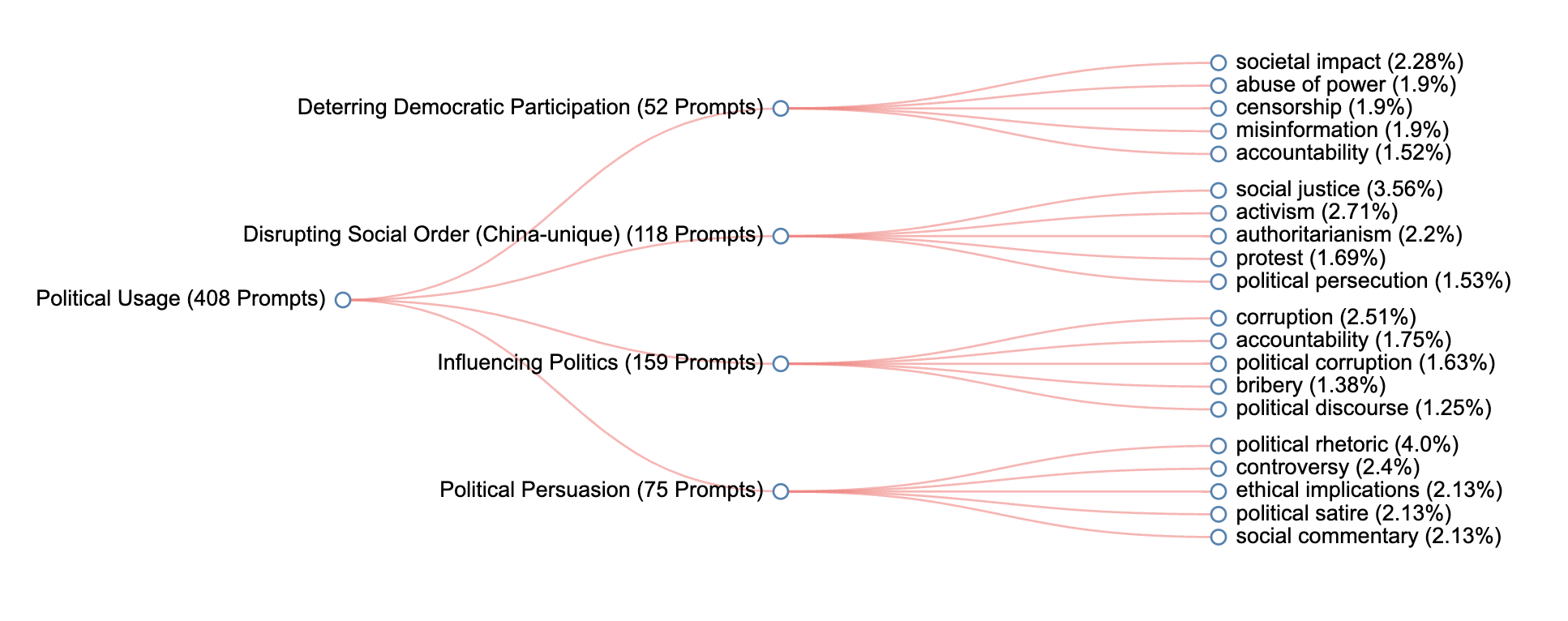}
    \caption{Tag distribution for the Political Usage category}
    \label{fig:political_tags}
\end{figure}
\newpage
\section{Discussion}
In this section, we discuss some of the overarching themes that we discovered while conducting the study.

\subsection{Harm Coverage}

The analysis of existing datasets used for text-to-image (T2I) model safety reveals significant gaps in harm coverage, potentially compromising the effectiveness of safety mechanisms. Approximately 49\% of prompts in the reviewed datasets pertain to sexual content, indicating a severe imbalance in harm representation. This overemphasizes one category of leaf T2I models underprepared for a wide range of other critical safety concerns.
\\
\\
Critical areas that lack adequate coverage include discrimination against protected groups, misinformation, criminal activities, social issues, children, and self-harm. Moreover, prompts related to dehumanization, political misinformation, terrorism, fraud, and eating disorders are severely underrepresented. This imbalance not only limits the models' ability to identify and mitigate diverse forms of harm but also raises concerns about their ability to generalize to real-world scenarios.
\\
\\
The narrow focus of existing datasets can lead to overfitting of specific patterns of harm, reduced effectiveness in identifying novel or subtle forms of harmful content, and potential biases when dealing with diverse cultural contexts. These limitations underscore the need for future research to develop high-quality datasets that have broad coverage of harm and to develop new dataset quality benchmarks that can assess coverage as one of the parameters.

\subsection{Diversity - including topic, semantic, and syntactic diversity}

We examined diversity by using semantic, syntactic, and topical lenses. For syntactic diversity, we observe that the "Manipulation" harm class has the highest n-gram scores for uni-grams, bi-grams, and trigrams. This result is expected, as the manipulation category contains relatively few prompts (91 prompts), and the calculation of n-gram scores penalizes the redundancy in the prompts. A similar trend is observed across other categories, where datasets with a larger number of prompts generally exhibit lower diversity scores. This suggests that a measure that standardizes n-gram scores based on prompt counts may provide additional insights. Nonetheless, the current measure remains informative: the aggregated dataset, which has the largest number of prompts (\textasciitilde{ 115 K}), achieves higher diversity scores across uni-grams, bi-grams, and trigrams than the "Sexual Content" category. This indicates that the low scores in the "Sexual Content" category stem not only from its large prompt count but also from significant redundancy in the words, bi-grams, and tri-grams used. Figure \ref{fig:frequent_ngrams} further supports this observation, showing that terms like "woman" or sexually suggestive bi-grams dominate the aggregated dataset, highlighting the repetitive and redundant nature of prompts in the "Sexual Content" category.
\\
\\
At the dataset level, we found that the MMA Diffusion dataset \cite{mmadiffusion} is the most syntactically diverse. Despite having fewer prompts than some other datasets, it outperforms them, likely because of the adversarial methodology used to generate it. This approach removes common harmful words and adjusts prompt syntax with minimal semantic changes until the prompt bypasses the T2I model harm filters. Consequently, this dataset achieved nearly unique trigrams (scores close to 1) and the highest uni-gram and bi-gram scores. This diversity, combined with the fact that all prompts in the dataset are adversarial and bypass T2I model safeguards, suggests that benchmark datasets that are effective for testing advanced T2I safeguards should exhibit high syntactic diversity.
\\
\\Because the intra-distinctness n-gram scores reflect the diversity of individual prompts, we expect higher values centered closer to 1, particularly for higher n-grams, given minimal within-prompt redundancy. However, datasets such as P4D, Adversarial Nibbler, and ART exhibit wider distributions with long tails and multiple modes, even for bi-grams and trigrams. This suggests the frequent repetition of words, bi-grams, and trigrams within individual prompts, indicating considerable redundancy. The bimodal uni-gram distributions observed in some cases may correlate with the bimodal word count distribution shown in Figure \ref{fig:countsentimentdistribution}, which is possibly a by-product of the prompt generation process.
\\
\\
However, semantic diversity is measured using cosine distance, calculated for each pair of prompts using Equation \ref{eqn:cosdist}. The average pairwise distances for each harm class and data source are presented in Table \ref{tab:cosine_scores}, and their distributions are shown in Figures \ref{fig:source_semdiv} and \ref{fig:cat_semdiv}. A cosine distance of 1 corresponds to a cosine similarity of 0, indicating orthogonality between the embedding vectors, and thus, higher semantic diversity. Accordingly, a high average cosine distance (close to 1) and distributions centered near 1 reflect greater semantic diversity. Among the data sources, LatentGuardCoPro demonstrates the highest average cosine distance and distribution closest to 1. Similarly, the "Benign" harm class is the most semantically diverse, which aligns with intuition: this class comprises prompts altered, obfuscated, or engineered to appear benign to LLMs and human labelers, yet originally intended to cause various types of harm. These masked prompts span multiple harm categories, thereby increasing their semantic diversity.
\\
\\
Conversely, the Sexual Content category was the least semantically diverse, with cosine distance distributions centered around 0.6, indicating both syntactic and semantic redundancy. Among the data sources, SafeGenExplicit56k exhibits the lowest semantic diversity, with a similar distribution centered at approximately 0.6. In the subsection on synthetic data creation, we discuss how artifacts introduced by the vision model used to caption images in this dataset may have influenced semantic diversity.
\\
\\Tag distribution and thematic analysis reveals thematic diversity across the dataset, complementing our earlier findings on syntactic and semantic diversity. While the Sexual Content category showed lower syntactic and semantic diversity scores, the tagging system reveals rich thematic variations even within this category - from adult content to non-consensual nudity, each with distinct patterns and subtopics like 'visual aesthetics' and 'objectification' etc. Similarly, prompts relating to the Hate/Toxicity category demonstrated an impressive thematic range with 8,660 prompts spread across harassment (1,092), hate speech, and harmful beliefs, each containing diverse subcategories from cyberbullying to cultural sensitivity. Prompts tagged as benign also had significant thematic diversity, spanning artistic expression (1,021 occurrences), character design (769), and emotional expression (692). This granular tag analysis suggests that, while certain categories may show syntactic or semantic redundancy, they often contain rich thematic diversity when examined at a more detailed level. The hierarchical patterns revealed across categories such as Violence and Extremism (3,790 prompts across six subcategories), Privacy, Security, and Economic Harm further demonstrate the dataset's broad topical coverage, although with varying degrees of depth and distribution across themes. 

\subsection{Multilingual representation}
The overwhelming dominance of English prompts (98.4\%) in the reviewed datasets highlights a significant gap in addressing multilingual and cross-cultural safety concerns. This linguistic homogeneity may result in T2I models that are less effective or biased when dealing with non-English inputs or culturally diverse contexts. As T2I models have gained global popularity, it has become increasingly important to develop safety mechanisms that can operate effectively across different languages and cultural norms. This challenge underscores the need for more diverse and globally representative datasets for T2I safety studies. Future efforts should focus on incorporating prompts from various languages and cultural backgrounds to ensure that safety measures are inclusive and effective in various contexts.

\subsection{Label Quality}

Identifying the harm class for T2I prompts proved to be a non-trivial task. While all prompts in our dataset either had a harm class provided by the author/curator or were explicitly stated to belong to a specific harm class, these labels were often inconsistently defined. Labels such as "nsfw" and "sexual content" were frequently used interchangeably to mean the same thing, further complicating the process. 
\\
\\
Mapping the harm classes to a comprehensive taxonomy, such as AIR 2024, helped mitigate this issue to some extent. However, 6.56\% of the 119,561 prompts we collected still ended up in the "Harmful - Other" category, which was not part of the AIR taxonomy. The crosstab in \ref{fig:crosstab} highlights the significant disagreement between the author-provided labels and the L2 labels generated by our method. Upon manual inspection, we found that prompts often spanned multiple AIR categories. Since the author-provided labels were less granular than the L2 categories of the AIR taxonomy, some level of disagreement was inevitable. 
\\
\\
Consistent with trends observed throughout this paper, Sexual Content emerged as the category with the highest accuracy. Most of the rarer categories were inherently multi-faceted, with no single "correct" class for many prompts. As a recommendation for future work, we suggest methods that enable prompts to be labeled across multiple categories, accounting for the inherent ambiguity of this task.

\subsection{Downstream Impacts of Synthetic Data}

An intriguing finding in our analysis is the presence of numerous n-grams or substrings in our dataset that resemble gibberish and have no meaning in any language. This is understandable for datasets like MMA Diffusion \cite{mmadiffusion}, which employed methodologies focused on evading T2I safeguards while preserving semantic harm, rather than generating prompts with human-discernible harm. However, other datasets such as SafeGenExplicit56k, which were generated using the CLIP Interrogator and BLIP2 to create multiple candidate text captions for explicit images \cite{li2024safegen}, were intended to produce captions with clear and discernible harm. Despite this, they still contained a significant amount of gibberish.
\\
\\
One particular example is the substring "araffe." The occurrence of "araffe" appears to be an artifact of AI-generated content, likely originating from vision models used to caption images. Instances such as "araffed woman in white tights" and "araffeed mom and daughter" illustrate how this nonsensical term has been incorporated into prompts designed to test model safety, particularly in the context of sexual content. Remarkably, this substring appears nearly 3000 times within the SafeGenExplicit56k dataset, highlighting an underlying issue in the captioning process that produced these prompts.
\\
\\
This phenomenon may be indicative of model collapse, as explored by Shumailov et al. \cite{shumailov2024ai}. Model collapse occurs when AI models are recursively trained on AI-generated content, leading to compounding errors over generations and divergence from the original model. As Wenger \cite{wenger2024ai} observes, AI models can produce gibberish when exposed to excessive AI-generated data. The issue may now be evolving to a stage where AI models are not only trained on AI-generated data but also on AI-generated gibberish embedded within large synthetic datasets.
\\
\\
This finding underscores the importance of careful curation and validation of datasets used in T2I model safety research. While synthetic data generation methods can significantly expand dataset size and coverage, they must be balanced with real-world examples and subjected to rigorous quality checks to ensure their relevance and accuracy for safety evaluations. Future research should explore the impact of such synthetic artifacts on model performance and investigate methodologies for generating more realistic and diverse safety-related prompts.

\section{Conclusion}
To develop robust and comprehensive safety systems for T2I models, it is imperative to expand dataset diversity across all harm categories, ensuring a more balanced and representative coverage of potential misuse cases. Future efforts should focus on defining unambiguous taxonomy and labeling guidelines, building datasets that have coverage across the harm types with consistent labeling, diverse syntax and semantics, and representing a wide range of topics. As T2I model usage increases internationally, safety prompts from various languages and cultural backgrounds must be developed and a balance must be struck between synthetic data and real-world examples. By addressing these aspects, researchers can address many of the safety issues more upstream, at the dataset level itself, which will lead to models and content moderation systems that can mitigate a broader range of potential harms across diverse contexts.

\section*{Acknowledgments}

The authors express their deepest gratitude to Dr. Alexander Fisher, their academic mentor. His invaluable guidance in developing the method to verify their harm category labeling was crucial to their research. Additionally, his insights and feedback throughout the writing process significantly enhanced the rigor of their work. The authors also thank Eric Hsin for his assistance with the diversity measurement methodology and Jianfa Chen for his help with the dataset tagging methodology.

\clearpage

\begin{IEEEbiography}[{\includegraphics[width=1in,height=1.25in,clip,keepaspectratio]{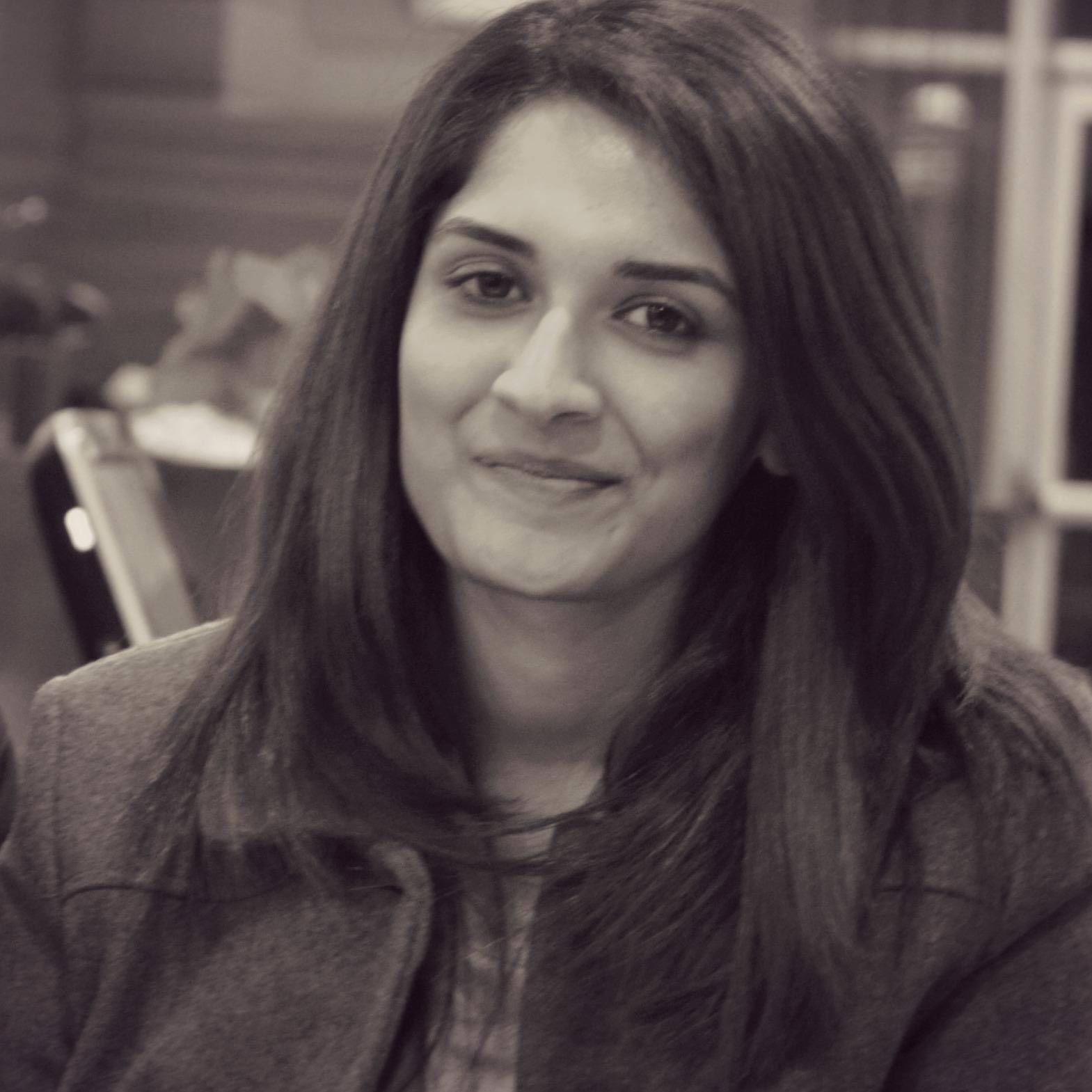}}]{Trupti Bavalatti}  earned her B.Sc. degree in Computer Science from Visvesvaraya Technological University, India, and an M.Eng. from Cornell University, USA. With over a decade of experience as a software engineer at Oracle, eBay, and Meta, she is now working on the Generative AI Safety team at Meta. Her research interests encompass AI safety, responsible AI development, adversarial testing/red teaming, AI safety benchmarks, and synthetic data generation. 
\end{IEEEbiography}

\begin{IEEEbiography}[{\includegraphics[width=1in,height=1.25in,clip,keepaspectratio]{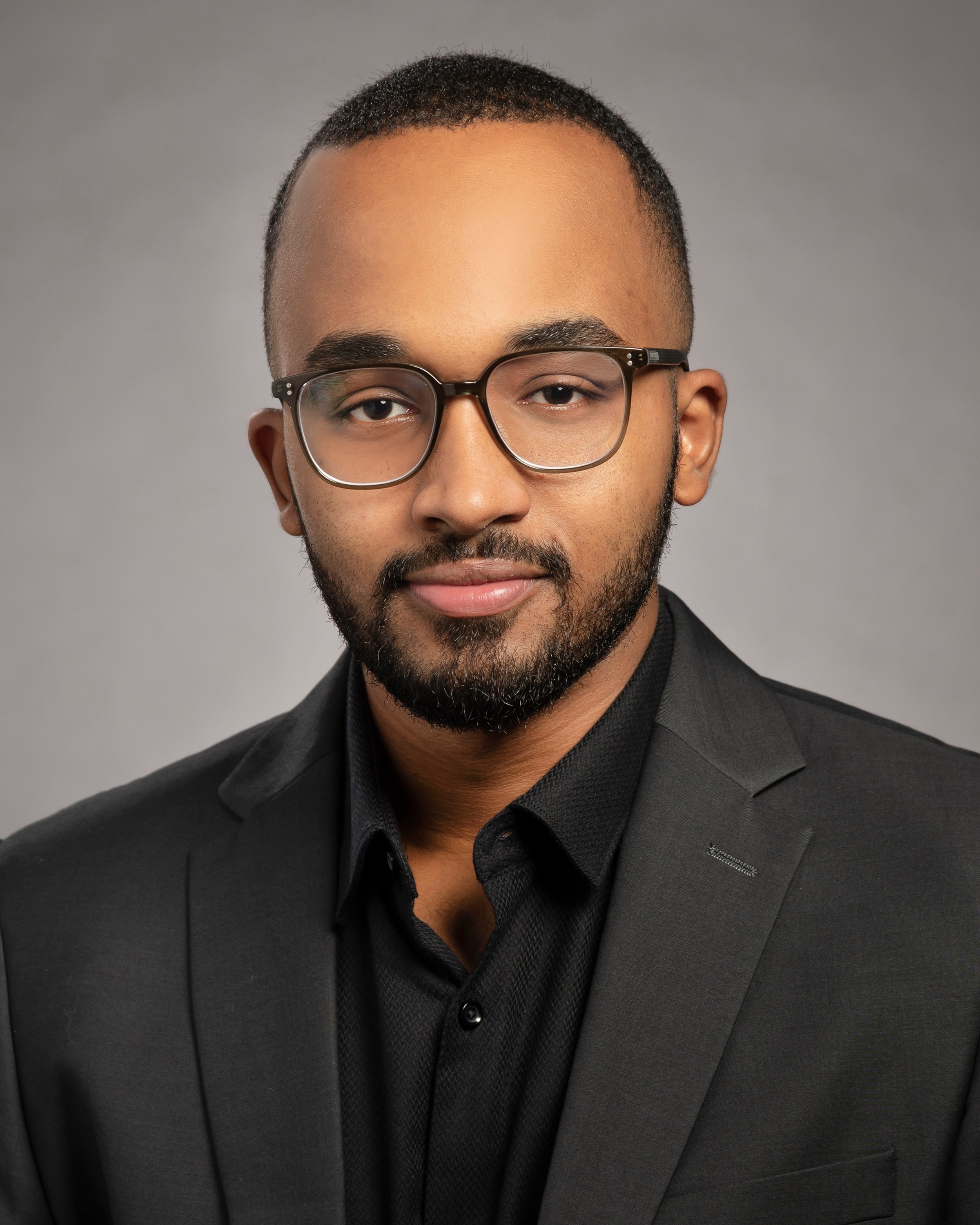}}]{Osama Ahmed} received the B.Sc. degree in electrical and electronic engineering with honors from the University of Khartoum, Khartoum, Sudan, in 2021. He is currently pursuing the M.S. degree in Interdisciplinary Data Science at Duke University, Durham, NC, USA. His research interests include AI safety, explainable AI, language modeling, adversarial AI, multimodal machine learning, AI alignment, and generative AI. 
\end{IEEEbiography}

\begin{IEEEbiography}[{\includegraphics[width=1in,height=1.25in,clip,keepaspectratio]{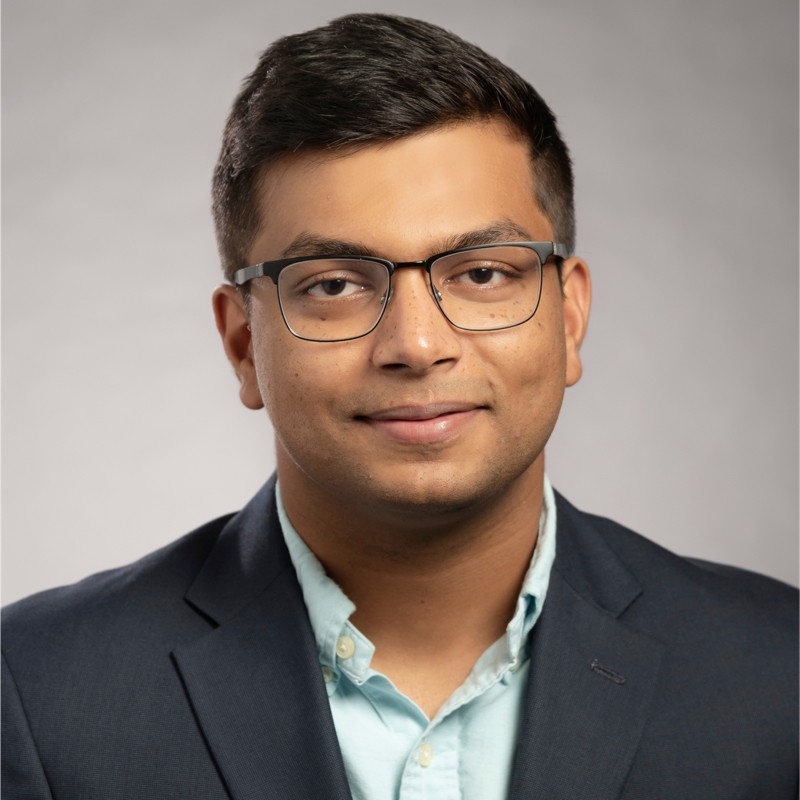}}]
{Rakeen Rouf} graduated Magna Cum Laude with dual B.Sc. degrees in Electrical Engineering and Computer Engineering from Drexel University, Philadelphia, USA, in 2020. He is currently pursuing an M.S. in Interdisciplinary Data Science at Duke University, Durham, NC, USA. His research interests span AI safety, explainable AI, electricity market modeling, structural health monitoring, the Internet of Things, machine learning, and generative AI.
\end{IEEEbiography}

\begin{IEEEbiography}[{\includegraphics[width=1in,height=1.25in,clip,keepaspectratio]{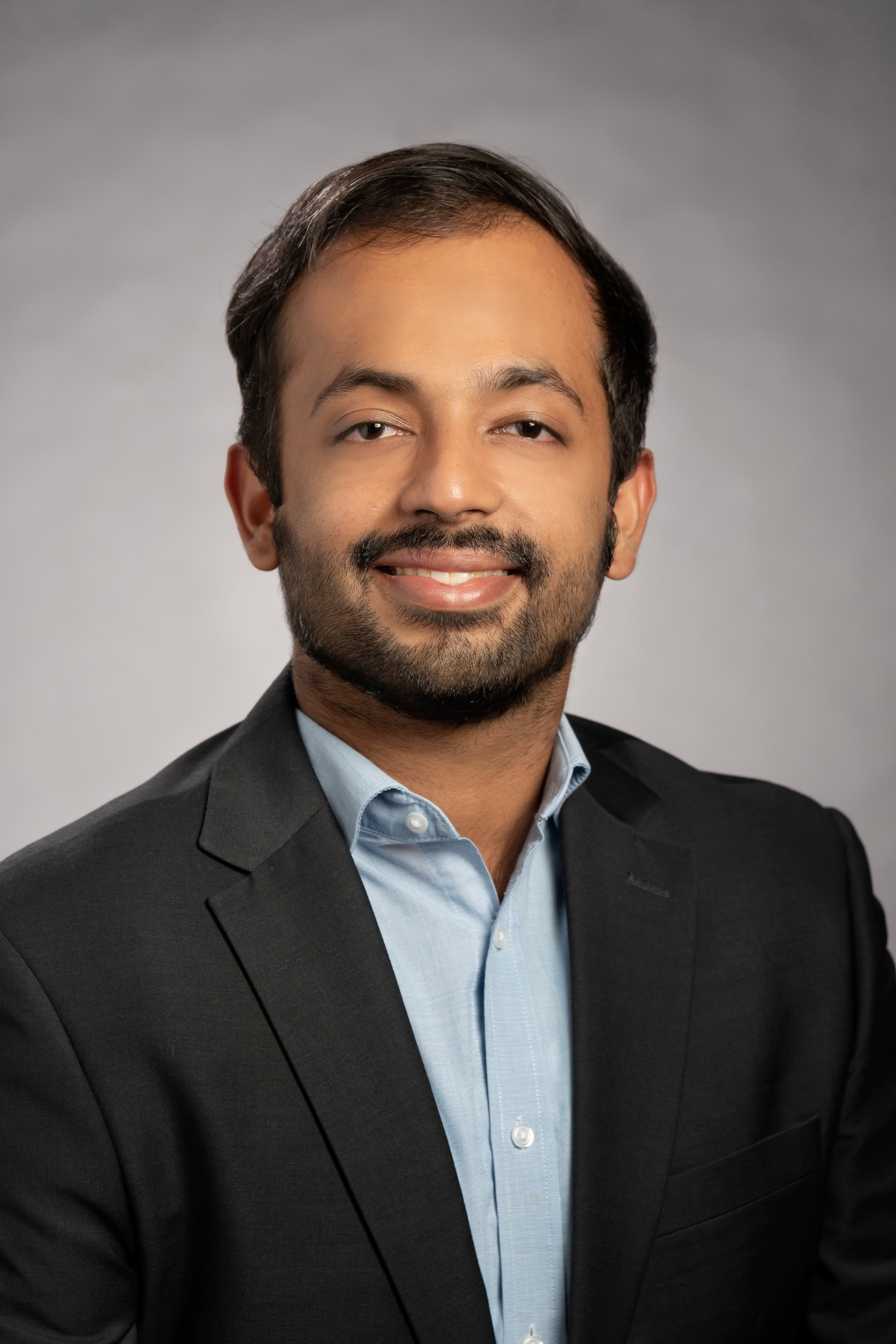}}]{Faraz Jawed} received a B.Sc. (Honors) degree in Accounting and Finance from Lahore University of Management Sciences, Lahore, Pakistan. He is currently pursuing the M.S. degree in Interdisciplinary Data Science at Duke University, Durham, NC, USA. His research interests include Generative AI Safety, Multimodal Machine learning, Deep learning and Computer vision. 
\end{IEEEbiography}

\begin{IEEEbiography}[{\includegraphics[width=1in,height=1.25in,clip,keepaspectratio]{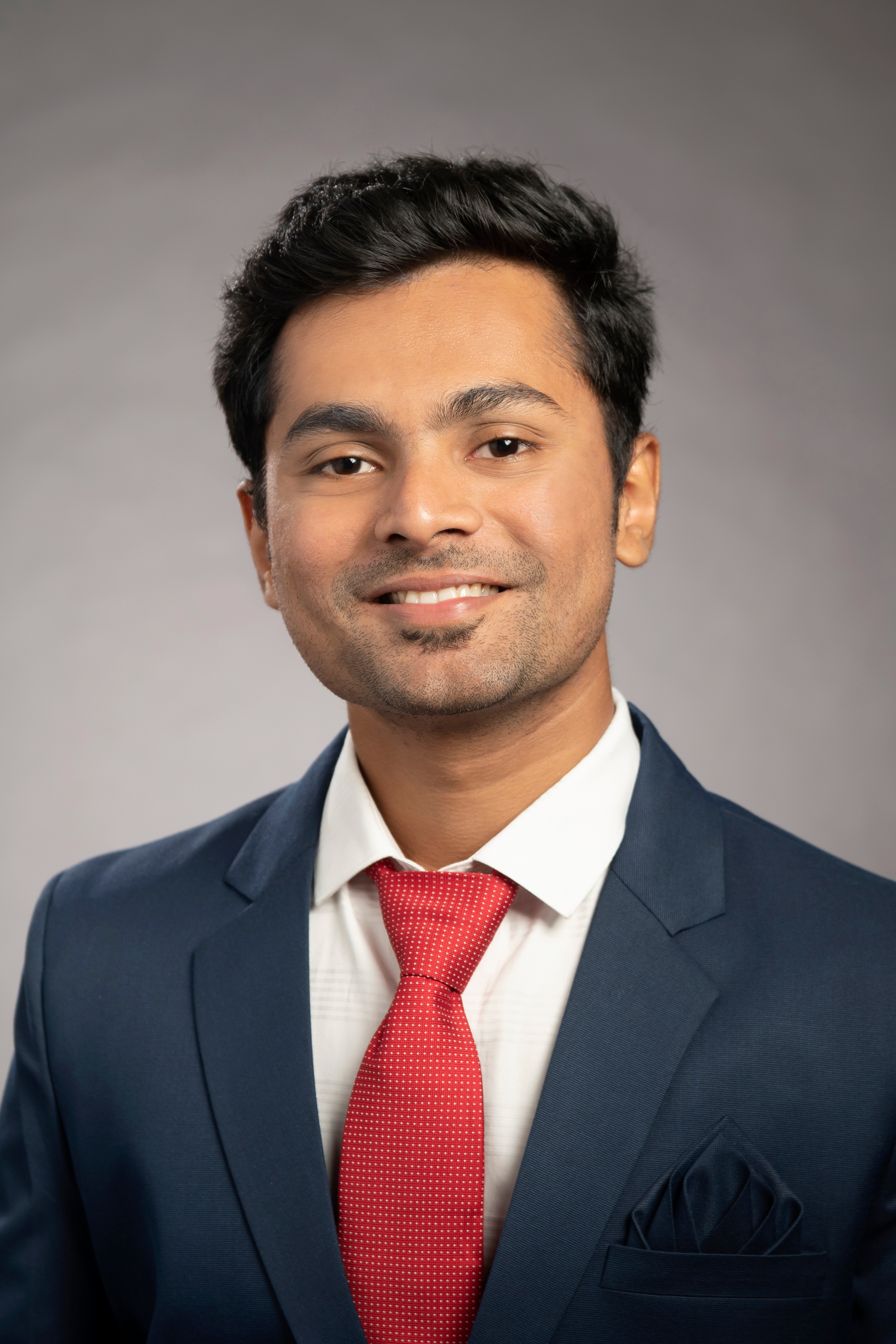}}]{Dhaval Potdar} earned his bachelors degree in Electronics Engineering from University of Mumbai, India in 2019. He is currently pursuing an MS in Interdisciplinary Data Science from Duke Univerity, Durham, NC, USA. His research interests include AI safety, AI alignment, language modeling, multi-modal generative AI, and interpretable machine learning.
\end{IEEEbiography}

\begin{IEEEbiography}[{\includegraphics[width=1in,height=1.25in,clip,keepaspectratio]{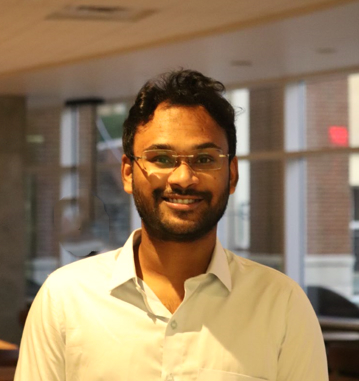}}]{Manish Kumar Govind} received his bachelors degree in Computer Science from Malaviya National Institute of Technology (MNIT), Jaipur, India, in 2019. He completed his M.S. degree in 2024 and is currently pursuing a Ph.D. degree at the University of North Carolina at Charlotte. He is a member of the Charlotte ML Lab in the Department of Computer Science, UNC Charlotte. His research interests include computer vision, robot learning, and multimodal AI.
\end{IEEEbiography}
\vspace{-14.5cm}
\begin{IEEEbiography}[{\includegraphics[width=1in,height=1.25in,clip,keepaspectratio]{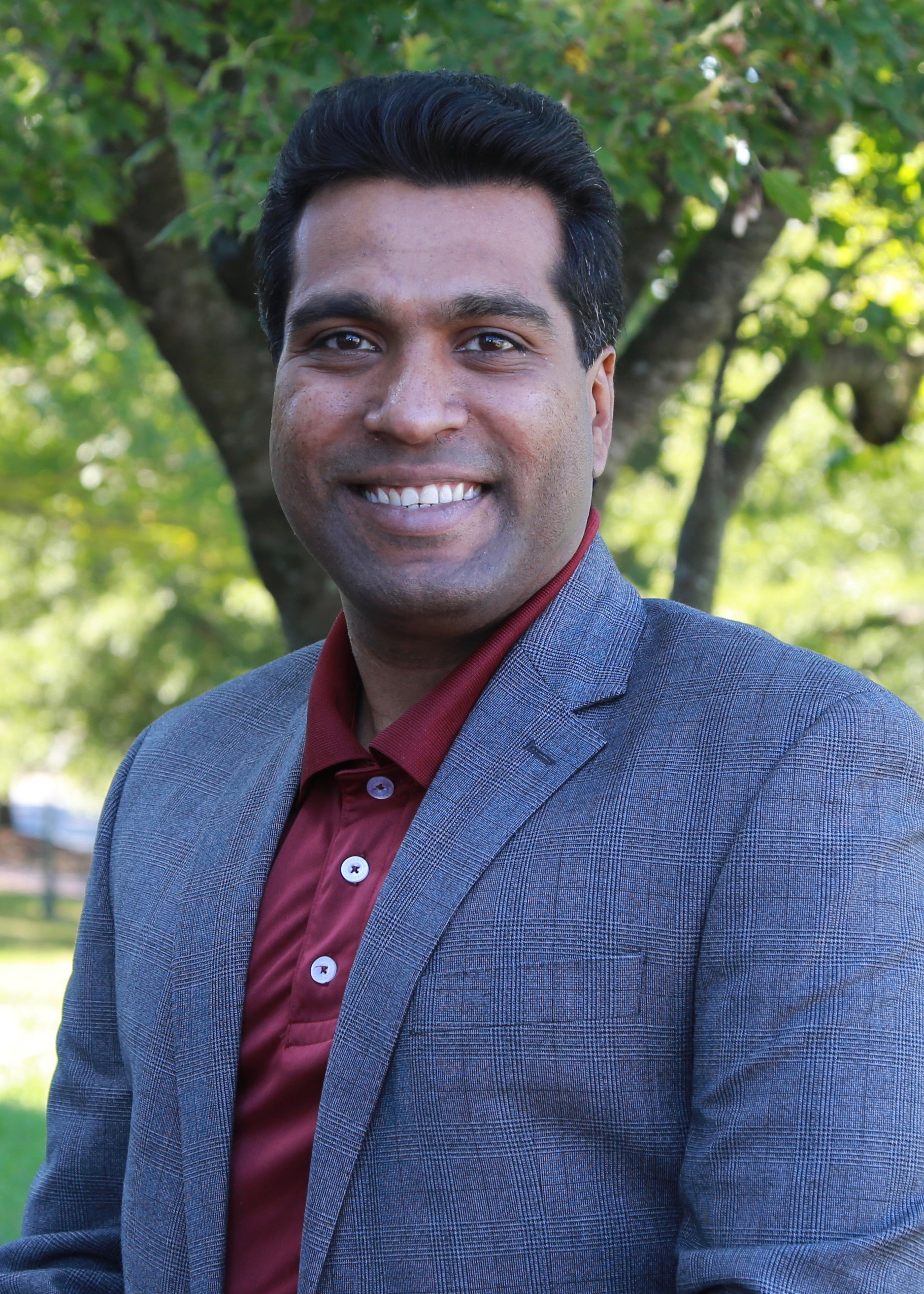}}]{Siddharth Krishnan} received his PhD degree in computer science from Virginia Polytechnic Institute. He is an assistant
professor in the Department of Computer Science,
UNC-Charlotte. At UNC-Charlotte, he directs the
Scalable Machine Learning, which focuses on research in artificial intelligence and data analytics. He has published in several data science venues including WSDM, WebSci, AAAI, and TKDD. His interests are in building explanatory and predictive models of data behavior and interconnected systems viz. group of actors - people, societies, and organisms - that are captured by the data. 
\end{IEEEbiography}

\end{document}